\documentclass[utf-8]{bytedance_seed}
\usepackage{CJKutf8}

\usepackage[toc,page,header]{appendix}

\usepackage{minitoc}
\usepackage{amsfonts}

\newcommand{\linyuan}[1]{\textcolor{blue}{#1}}

\title{Memory Retrieval and Consolidation in Large Language Models through Function Tokens}

\author[]{Shaohua Zhang}
\author[]{Yuan Lin}
\author[]{Hang Li}

\affiliation[]{ByteDance Seed}

\abstract{
The remarkable success of large language models (LLMs) stems from their ability to consolidate vast amounts of knowledge into the memory during pre-training and to retrieve it from the memory during inference, enabling advanced capabilities such as knowledge memorization, instruction-following and reasoning. However, the mechanisms of memory retrieval and consolidation in LLMs remain poorly understood. In this paper, we propose the function token hypothesis to explain the workings of LLMs: During inference, function tokens activate the most predictive features from context and govern next token prediction (memory retrieval).
During pre-training, predicting the next tokens (usually content tokens) that follow function tokens increases the number of learned features of LLMs and updates the model parameters (memory consolidation). Function tokens here roughly correspond to function words in linguistics, including punctuation marks, articles, prepositions, and conjunctions, in contrast to content tokens. We provide extensive experimental evidence supporting this hypothesis. Using bipartite graph analysis, we show that a small number of function tokens activate the majority of features. Case studies further reveal how function tokens activate the most predictive features from context to direct next token prediction. We also find that during pre-training, the training loss is dominated by predicting the next content tokens following function tokens, which forces the function tokens to select the most predictive features from context.
}

\date{\today}
\correspondence{\email{\{zhangshaohua.cola,linyuan.0,lihang.lh\}@bytedance.com}}

\usepackage{longtable}
\usepackage{booktabs}
\usepackage{multirow}  
\begin{document}
\maketitle


\section{Introduction}
Large Language Models (LLMs)~\citep{openai2022chatgpt,NEURIPS2020_1457c0d6,openai2023gpt,anthropic2024claude,team2023gemini,liu2024deepseek} have demonstrated remarkable capabilities. They possess strong knowledge memorization abilities, ranging from remembering simple factual knowledge (e.g., \textit{The capital of the United States is Washington, D.C.}) to the verbatim reproduction of lengthy passages (e.g., \textit{Recite Martin Luther King Jr's ``I Have a Dream" speech word by word}). Beyond that, LLMs also exhibit strong general skills, such as instruction following~\citep{ouyang2022training,wei2021finetuned} (e.g., \textit{As a financial analyst: explain quantitative tightening, then list three stock market impacts.}) and reasoning~\citep{wei2022chain,kojima2022large} (e.g., \textit{The streets are wet and the sidewalks are slick. What is the most likely explanation?}). 

In the human brain, long-term memory forms through synaptic consolidation, where the synapses between neurons are strengthened, ultimately creating neural circuits that store knowledge~\citep{josselyn2020memory}. Inspired by this biological mechanism, artificial neural networks have been developed. These systems consist of neurons linked by weighted connections, and their weights (parameters) are obtained by training on data. The weights of a neuron determines how it responds to its inputs to produce an activation~\citep{geva2021transformerfeedforwardlayerskeyvalue}. 
A technique utilizing Sparse Autoencoders (SAEs)~\citep{cunningham2023sparseautoencodershighlyinterpretable} has been developed recently to analyze Transformer-based LLMs~\citep{vaswani2017attention}. It enables the decomposition of neuron activations into interpretable features, providing insights into how the circuits within the Transformer's layers are composed of these interpretable features~\citep{elhage2022superposition,chen2025personavectorsmonitoringcontrolling,hendel2023context}.

Despite significant progress in understanding LLM neuron activations, the memory mechanisms remain poorly understood. In particular, two fundamental questions are still not well addressed: (1) How is the memory retrieved during inference? and (2) How is the memory consolidated during pre-training? In this paper, we present our investigation into these questions. We find that analyzing from the perspective of function tokens and content tokens can help unravel the mystery of memory retrieval and memory consolidation.

In linguistics, function words are words that have little semantic meanings but play crucial grammatical and connective roles within and between sentences, such as articles, prepositions, and conjunctions~\citep{carnap2014logical}.
In contrast, content words are words that convey semantically explicit and rich meanings. The distribution of words in natural language follows Zipf's law~\citep{kanwal2017zipf}. In this distribution, function words occur with disproportionately high frequencies, occupying the head, while content words appear with much lower frequencies, forming the long tail. LLMs utilize tokens, which may represent words, sub-words, or punctuation marks. In our work, for ease of experimentation,
we automatically classify tokens into `function tokens' and `content tokens' based on their frequencies in the pre-training corpus, using this as an approximation of the linguistic concepts. 

\begin{figure}
    \centering
    \includegraphics[width=1.0\linewidth]{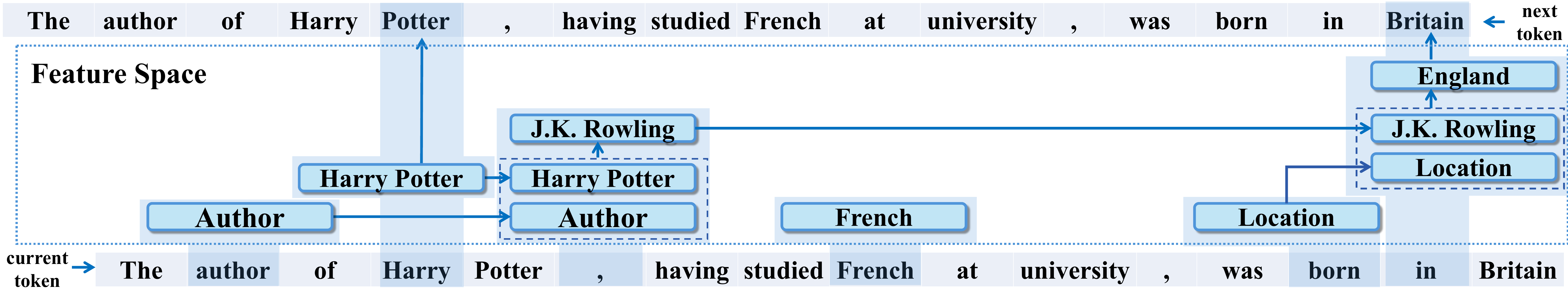}
    \caption{Function tokens can dynamically activate the most predictive features from the context to guide the next-token prediction. For example, the function token `in' reactivates features `J.K. Rowling' and `Location' from context (while suppressing feature 'French') and activates `England' to predict `Britain'. In contrast, the content token `Harry' activates feature `Harry Potter'.}
    \label{fig:Illustration}
\end{figure}
To investigate the role of function tokens during inference, we construct a bipartite graph connecting tokens to features obtained via SAE decomposition. We show that, although few in number, function tokens activate a large proportion of the LLM's features. 
Furthermore, our case studies show that the activation patterns for function tokens differ from those for content tokens. Function tokens dynamically reactivate predictive features from the context, whereas content tokens show little evidence of this effect. To understand why feature activations are centered on function tokens, we conduct pre-training experiments. We track next-token prediction loss across four categories based on whether the current token and the next token are function or content tokens. We find that LLMs first learn to predict function tokens before gradually learning to predict content tokens, a process accompanied by an increase in the number of features and the learning of the parameters. Furthermore, pre-training is dominated by the prediction of content tokens that follow function tokens. These observations reveal why function tokens can access a large portion of the LLM's features. Based on these findings, we propose the Function Token Hypothesis (see an example in Figure~\ref{fig:Illustration}).

In this paper, the LLMs are GPT-type models with a Transformer decoder architecture, obtained through pre-training and post-training (including SFT and RL)~\citep{nakano2021webgpt,radford2018improving}. Both pre-training and inference are conducted autoregressively via next-token prediction. At each layer of the Transformer, a vector of activations (after the add-norm operation of FFN) can be created, with each dimension representing a neuron. SAE can be performed on this activation vector to obtain a linear combination of features for each neuron. Here, knowledge refers to the LLM's parameters as well as all possible features that can be derived from them. Memory is the virtual system that stores the knowledge. Memory retrieval means the activations of features and circuits~\citep{olah2020zoom,elhage2021mathematical,wanginterpretability,merullocircuit}, while memory consolidation means the learning of the parameters to form and expand features and circuits.

\noindent\textbf{Function Token Hypothesis.} During inference, \textit{function tokens activate the most predictive features from the context to direct the next-token prediction} (memory retrieval). During pre-training, \textit{predicting content tokens based on the function tokens} drives the LLM to update its parameters to learn and expand features (memory consolidation).

The function token hypothesis is also supported by many phenomena observed in LLM research. For example, activations with unusually large magnitudes often occur at the initial tokens, periods, or newlines~\citep{sun2024massiveactivationslargelanguage}. Meaningless separator tokens disproportionately affect attention compared to semantically rich tokens~\citep{chen2025sepllmacceleratelargelanguage}. The use of `pivot tokens' during post-training can significantly enhance performance in response~\citep{abdin2024phi}. Training that concentrates on high-entropy tokens also yields better performance~\citep{wang20258020rulehighentropyminority}. We argue that these tokens are all function tokens that behave as the hypothesis predicts.

We believe that unraveling the important role of function tokens in LLM memory mechanisms not only enhances research on LLM interpretability but also provides insights for designing advanced learning algorithms, particularly those for enhancing alignment with human values.

The main contributions of this paper are summarized as follows: 
\begin{itemize}
    \item We demonstrate that during inference, function tokens are responsible for activating the most predictive features from the context to govern next-token prediction.

    \item We show that feature growth during pre-training is driven by the prediction of content tokens that follow function tokens.
    
    \item We propose the Function Token Hypothesis for explaining LLM memory mechanisms.
\end{itemize}

\section{Preliminary}
\subsection{Model Memory and Superposition Phenomenon}

\noindent\textbf{Feed-Forward Network as Key-Value Memory} Existing work views the Feed-Forward Network (FFN) layer in each block of a Transformer as a key-value memory or a neural memory~\citep{geva2021transformerfeedforwardlayerskeyvalue}. Specifically, the FFN can be formulated as (bias terms are omitted, as in common practice):
\begin{align}
&\textbf{z} = \text{ReLU}(\textbf{x} \cdot \mathbf{W}_{k}^\top) \\
&\textbf{y} = \textbf{z}\cdot \mathbf{W}_{v} \end{align}

Here, $\textbf{x}\in\mathbb{R}^d$ is the input vector, $\textbf{y}\in\mathbb{R}^d$ is the output vector, $\textbf{z}\in\mathbb{R}^{d_m}$ is the weight vector, $\mathbf{W}_k\in\mathbb{R}^{d_m\times d}$ is the key matrix, $\mathbf{W}_v\in\mathbb{R}^{d_m\times d}$ is the value matrix, and $d_m$ denotes the memory size. The output vector $\textbf{y}$ is in fact the activation of the FFN layer, where each dimension corresponds to a neuron. 

In the key-value memory interpretation, there are $d_m$ pairs of key vector and value vector. Each row of $\mathbf{W}_k\in\mathbb{R}^{d_m\times d}$ corresponds to a key vector $\textbf{k}_i\in\mathbb{R}^d$ and each row of $\mathbf{W}_v\in\mathbb{R}^{d_m\times d}$ corresponds to a value vector $\textbf{v}_i\in\mathbb{R}^d$. Given the input vector $\textbf{x}$, the similarity between $\textbf{x}$ and each of the key vectors $\textbf{k}_i$ is first calculated as $z_i=\text{ReLU}(\textbf{x}\cdot \textbf{k}_i^\top) \ge 0$, where ReLU acts as an unnormalized weighting function; the weighted sum of the corresponding value vectors based on the similarities is then calculated and output as $\textbf{y} = \sum_{i=1}^{d_m} z_i \mathbf{v}_i$. The interpretation suggests that knowledge of the Transformer is represented in the parameters of the FFN layers.
Note that Transformer attention layers also form key-value memories using softmax weighting.

\noindent\textbf{Superposition Phenomenon} Recent work on LLM interpretability shows the phenomenon of superposition~\citep{elhage2022superposition}, in which features can be extracted from the activations of neurons in a Transformer-based LLM.
The number of extracted features usually far exceeds the number of neurons. There exist many polysemantic neurons, each of which represents multiple meanings. 

Through sparse dictionary learning, the activations of polysemantic neurons can be decomposed into monosemantic features, each corresponding to a distinct, human-interpretable concept, such as the Golden Gate Bridge~\citep{templeton2024scaling}. 
A widely used method for dictionary learning is the Sparse Autoencoder (SAE), which learns to linearly decompose neuron activations through a reconstruction task. SAE decomposes an activation $\mathbf{y}\in\mathbb{R}^{d}$, typically the output of a specific layer, into a linear combination of features:
\begin{equation}
\mathbf{y} = \sum_{i=1}^{n} c_i \mathbf{f}_i = c_1 \mathbf{f}_1 + c_2 \mathbf{f}_2 + ... + c_n \mathbf{f}_n.
\end{equation}
Here, $\textbf{x}\in\mathbb{R}^d$ is the input vector, $\textbf{y}\in\mathbb{R}^d$ is the output vector, $\textbf{z}\in\mathbb{R}^{d_m}$ is the weight vector, $\mathbf{W}_k\in\mathbb{R}^{d_m\times d}$ is the key matrix, $\mathbf{W}_v\in\mathbb{R}^{d_m\times d}$ is the value matrix, and $d_m$ is the memory size. The output vector $\textbf{y}$ is the activation of the FFN layer, where each dimension corresponds to a neuron. 
Furthermore, the behavior of the LLM during generation can be partially controlled by steering the activations of features. For example, steering can control both specific concepts (e.g., Golden Gate Bridge~\citep{templeton2024scaling}) and behavioral patterns (e.g., sycophantic behavior~\citep{panickssery2023steering}). Similar feature activation phenomena are observed in human memory recall, with empirical evidence supporting the existence of neurons representing either specific or general concepts.

\subsection{Function Tokens and Content Tokens}

\begin{figure}[htbp]
    \centering
    \begin{subfigure}[b]{0.49\textwidth}
        \centering
        \includegraphics[width=\textwidth]{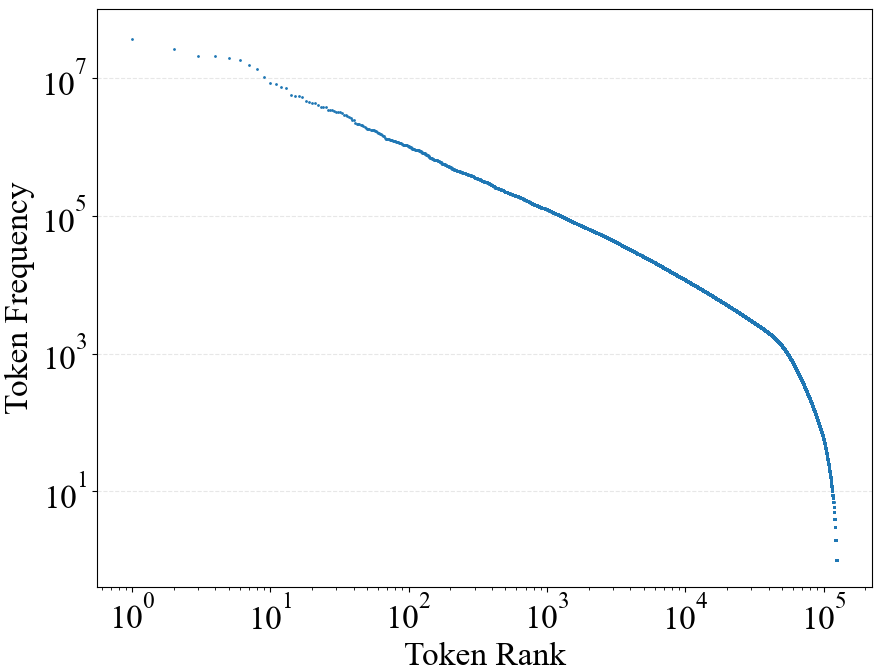}
        \caption{Zip'f distribution of tokens on a log-log scale.}
        \label{fig:token_sub1}
    \end{subfigure}
    \hfill
    \begin{subfigure}[b]{0.49\textwidth}
        \centering
        \includegraphics[width=\textwidth]{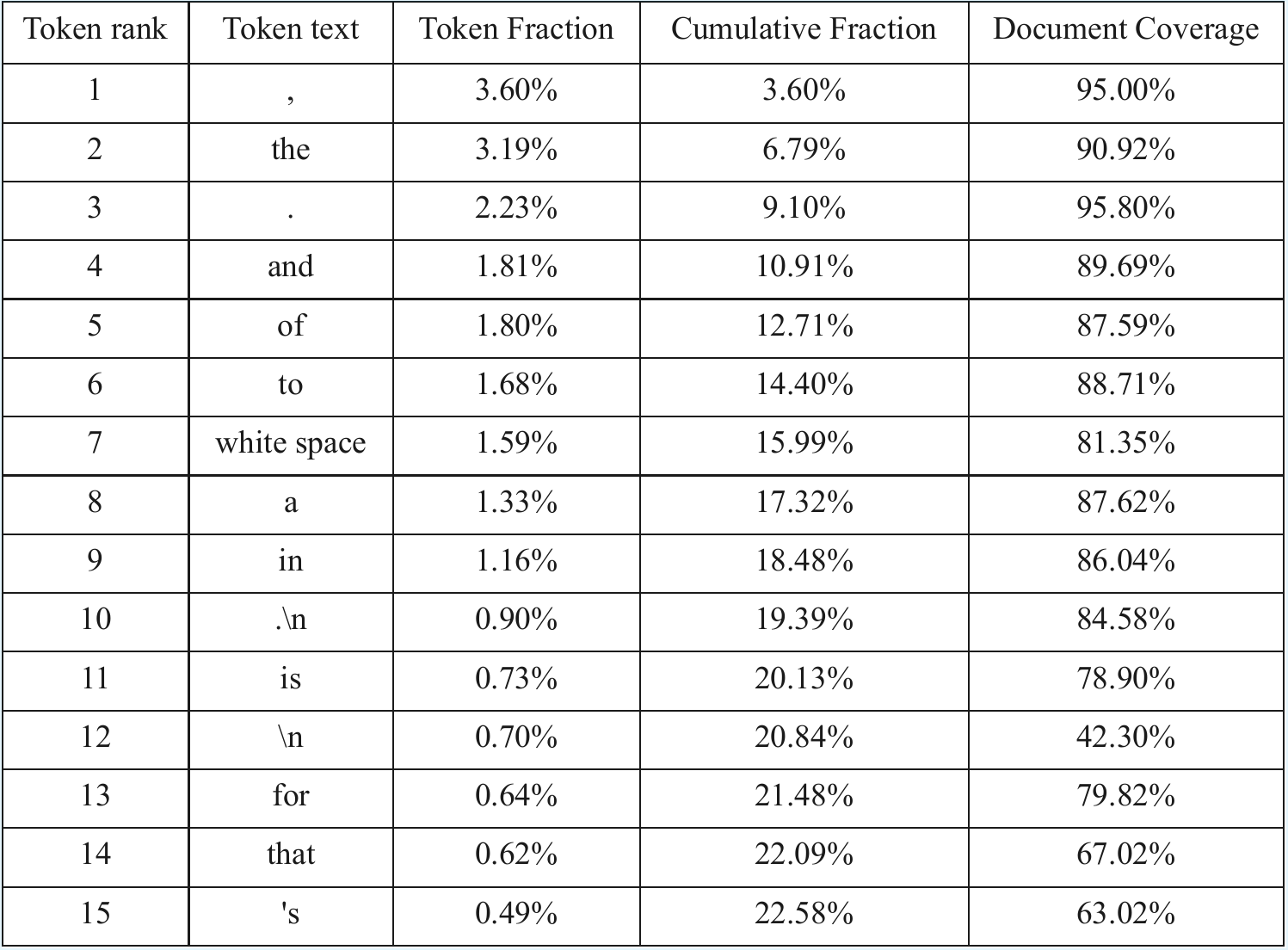}
        \caption{The 15 most frequent tokens.}\label{fig:top_tokens}
        \label{fig:token_sub2}
    \end{subfigure}
    \caption{Token frequency statistics in SlimPajama-627B.}
    \label{fig:main}
\end{figure}

We tokenized the SlimPajama-627B corpus~\citep{cerebras2023slimpajama}, a widely used pre-training dataset, using the LLaMA-3.1 tokenizer and sampled 1 billion tokens for statistical analysis. 
We group the tokens into function tokens and content tokens based on their frequency. This leverages the linguistic fact that function words typically have higher frequency, while content words have lower frequency. Starting from the most frequent, we add tokens until the set covered 40\% of all token occurrences, yielding 122 tokens labeled as function tokens; the rest are taken as content tokens. 
The resulting set of function tokens roughly corresponds to the function words defined in linguistics, with several exceptions like punctuation marks. The full list of function tokens appears in Appendix~\ref{apppend:func}.

\noindent\textbf{Token Frequency and Zipf's Law} As shown in Figure~\ref{fig:token_sub1}, token frequency follows the Zipf's law~\citep{piantadosi2014zipf}: $f(r) \propto r^{-\alpha}$, where $f(r)$ is the frequency of the token ranked $r$, revealing a fundamental property of natural language: a few tokens are used frequently, while most are used infrequently.  For example, the 15 most frequent tokens account for 22.58\% of the corpus (Figure~\ref{fig:token_sub2}).

\begin{figure}[htbp]
    \centering
    \begin{subfigure}[b]{0.32\textwidth}
        \centering
        \includegraphics[width=\textwidth]{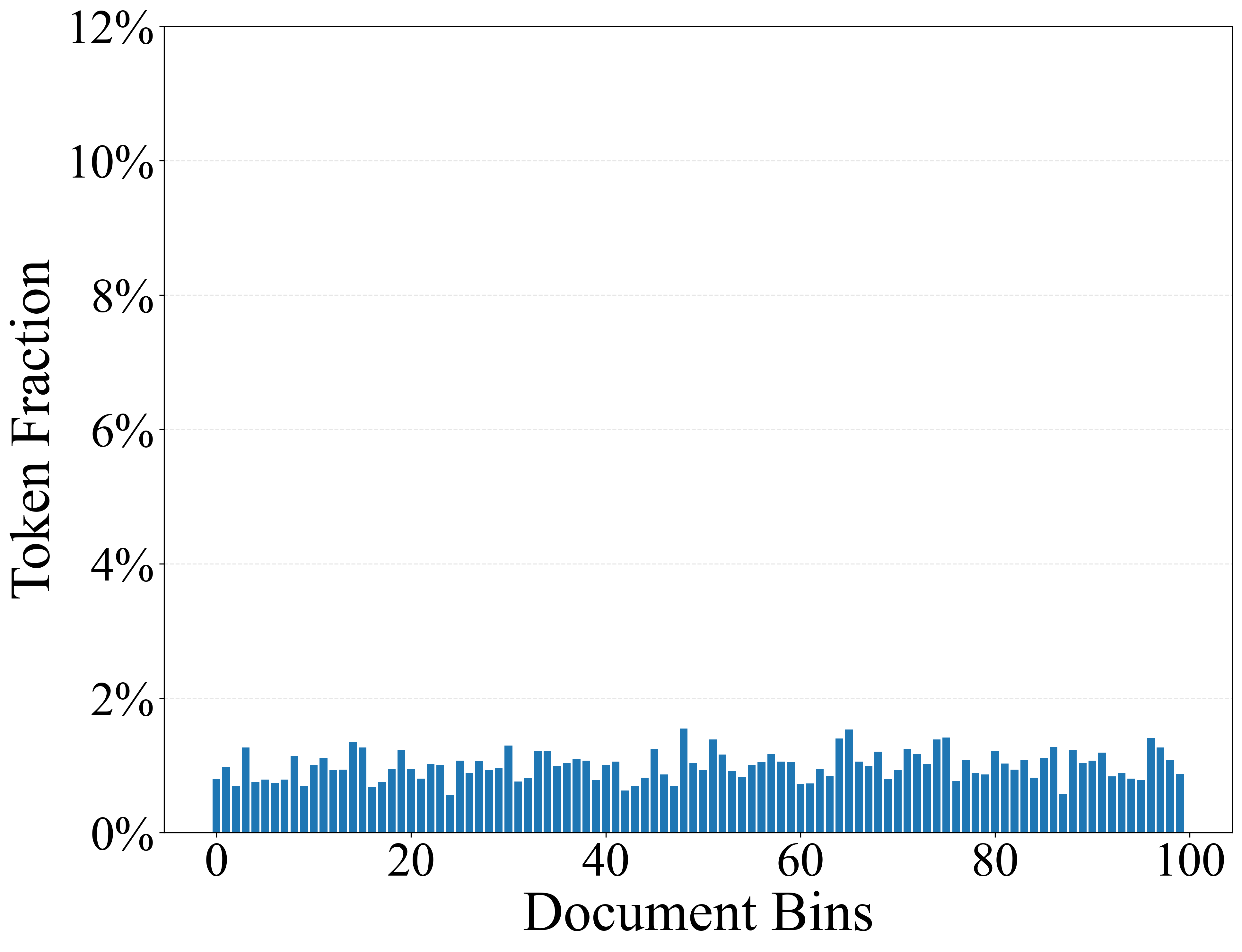}
        \caption{Distribution of the function token `of' across documents, showing uniform and dense coverage.}
        \label{fig:fig2_bursty_token_'of'}
    \end{subfigure}
    \hfill
    \begin{subfigure}[b]{0.32\textwidth}
        \centering
        \includegraphics[width=\textwidth]{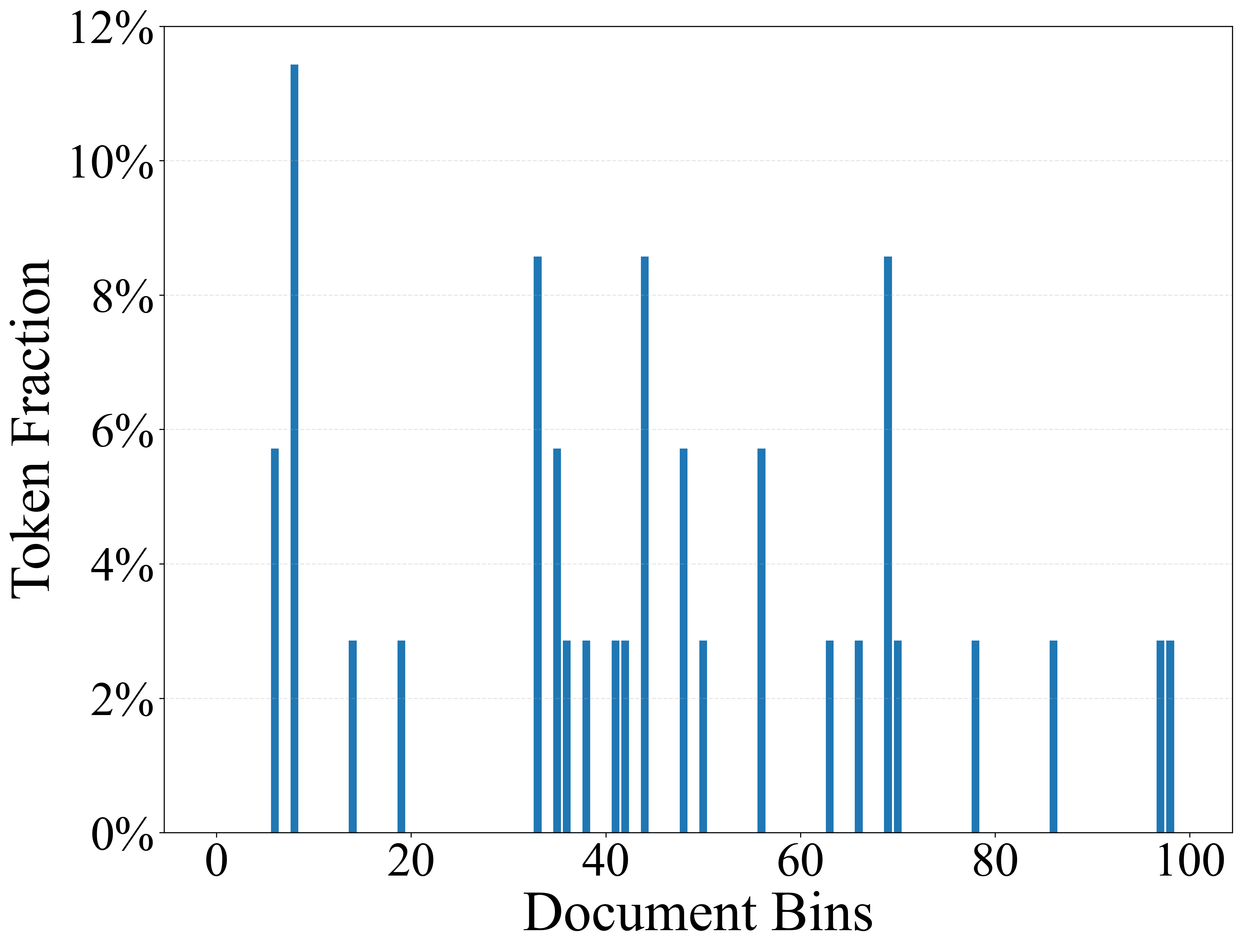}
        \caption{Distribution of the content token `Tokyo' across documents, showing sparse coverage.}
        \label{fig:fig2_bursty_token_'Tokeyo'}
    \end{subfigure}
    \hfill
    \begin{subfigure}[b]{0.32\textwidth}
        \centering
        \includegraphics[width=\textwidth]{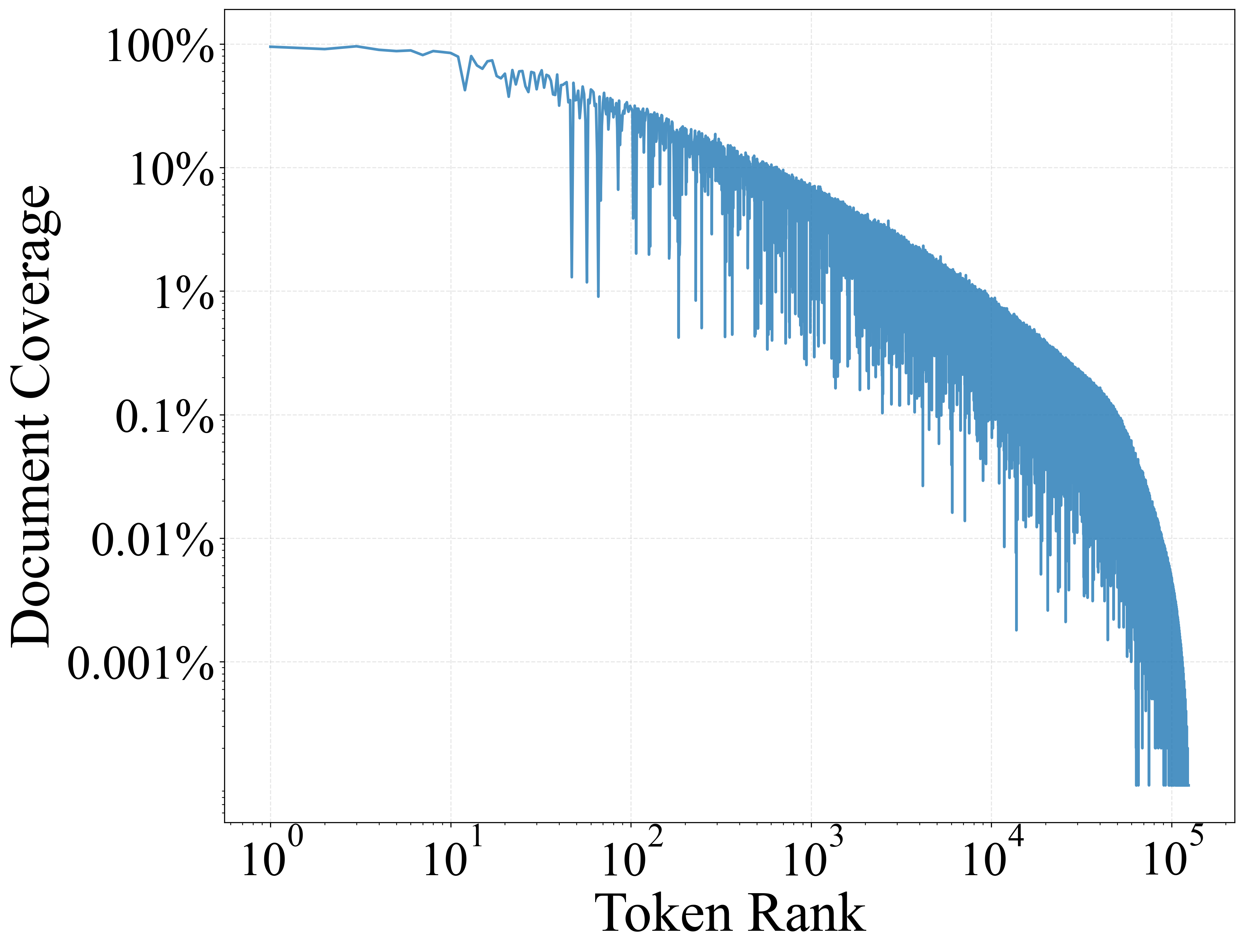}
        \caption{Document coverage versus token rank (ordered by frequency, log-log scale).}
        \label{fig:document_coverage}
    \end{subfigure}
    \caption{Distribution of function and content tokens. Document bins represent equal partitions of corpus documents.}
    \label{fig:bursty}
\end{figure}

\noindent\textbf{Document Coverage} A pre-training corpus contains a vast number of documents. High-frequency tokens are distributed uniformly across documents, while low-frequency tokens appear frequently within a limited number of documents~\citep{rychly2011words}, showing bursty distributions. For instance, as shown in Figure~\ref{fig:bursty}, the function token `of' appears with similar frequency across documents, whereas the content token `Tokyo' occurs only in a few. 
Thus, high-frequency tokens are utilized in nearly all training examples, while low-frequency tokens are used only in a small fraction of them. 


Figure~\ref{fig:document_coverage} shows a strong correlation between token frequency and document coverage: high-frequency tokens typically appear across most documents. Figure~\ref{fig:top_tokens} presents the 15 most frequent tokens, along with their corresponding document coverage values. 
Here the document coverage of a token $t$ is defined as $\frac{|\{d \in D : t \in d\}|}{|D|}$, 
where $D$ denotes the entire set of documents.

\section{Memory Retrieval through Function Tokens}
We study the relationships between tokens and model features during inference. 
The results show that a small set of function tokens can activate most features. Our case study reveals how the same function tokens create different activation patterns in different contexts, leading to different outputs. 

\begin{figure}
    \centering
    \includegraphics[width=1\linewidth]{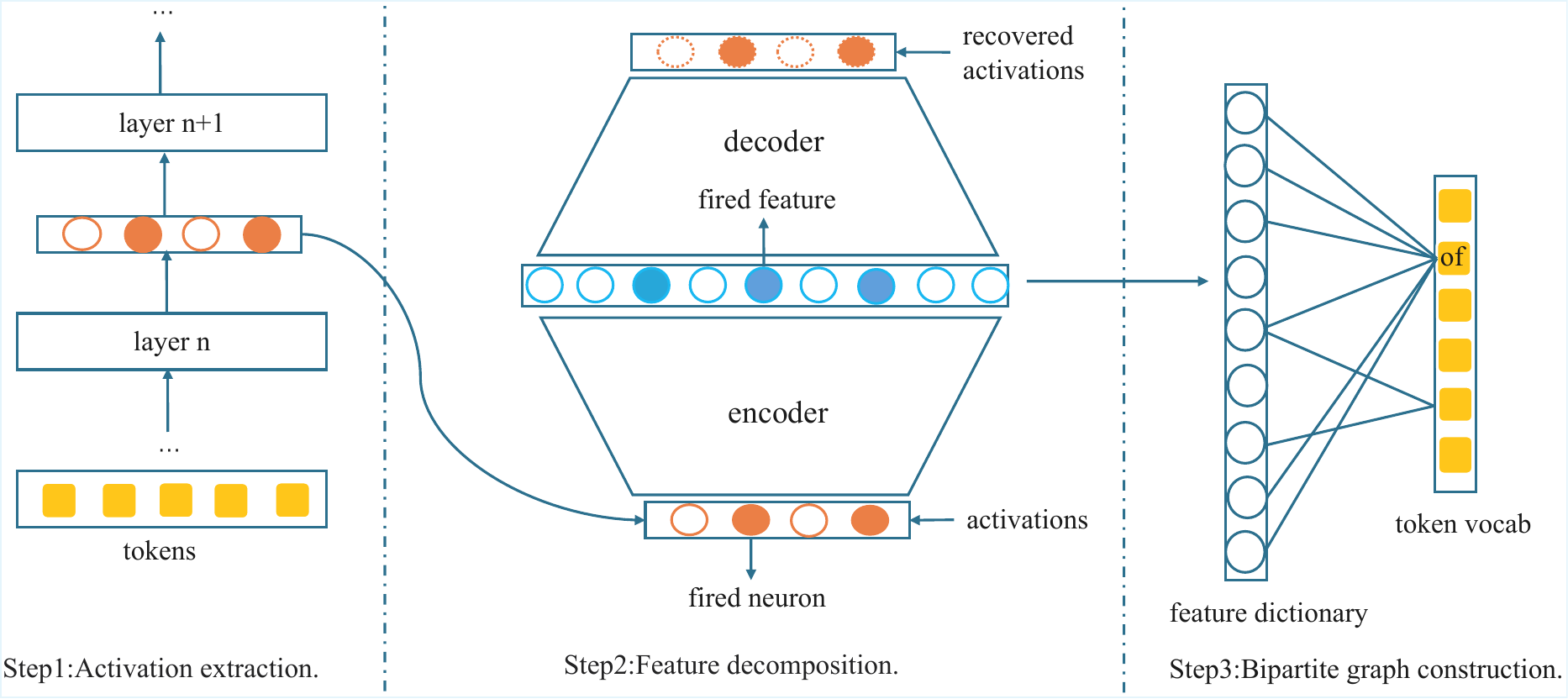}
    \caption{Construction of the bipartite graph using token-feature activation pairs as edges. Nodes consist of tokens from the vocabulary and features from the SAE decomposition.}
    \label{fig:Bipartite}
\end{figure}

\subsection{A Few Function Tokens Activate Most Features}


We use Gemma2-9B~\citep{gemmateam2024gemma2improvingopen} for our analysis, as it provides both models of different sizes and open-source SAEs~\citep{lieberum2024gemmascopeopensparse}. Gemma Scope has SAEs with varying dictionary widths. Among these, we select the SAE with the largest dictionary width, $2^{20}$,  to facilitate a more comprehensive feature decomposition.

To study how features are activated during inference, as illustrated in Figure~\ref{fig:Bipartite}, we construct a token-feature bipartite graph through the following steps:
\begin{itemize}
    \item Step 1:Extract activations. We feed 10,000 randomly sampled raw documents from the SlimPajama validation dataset into Gemma2-9B, with approximately 5 million tokens, and extract activations from the residual stream. We focus on three representative layers: layer 9 (shallow), layer 20 (middle), and layer 31 (deep).
    \item Step 2: Feature decomposition.For each layer, we apply the corresponding SAE to decompose token activations into sparse features. 
    \item Step 3: Bipartite graph construction. A token is linked to a feature if it activates the feature in a context. Each token-feature pair is connected by at most one edge, regardless of how many times the activation occurs. 
\end{itemize}
The bipartite graph comprises two node types: tokens and features. The number of token nodes equals the vocabulary size, while the number of feature nodes (each connected to at least one token) is 965,635, 947,341, and 919,220 for the three layers, respectively. With a dictionary width of $2^{20}$, this yields activation rates of 92.1\%, 90.3\% and 87.7\%, confirming sufficient coverage for analysis.


\begin{figure}[htbp]
    \centering
    \begin{subfigure}[b]{0.32\textwidth}
        \centering
        \includegraphics[width=\textwidth]{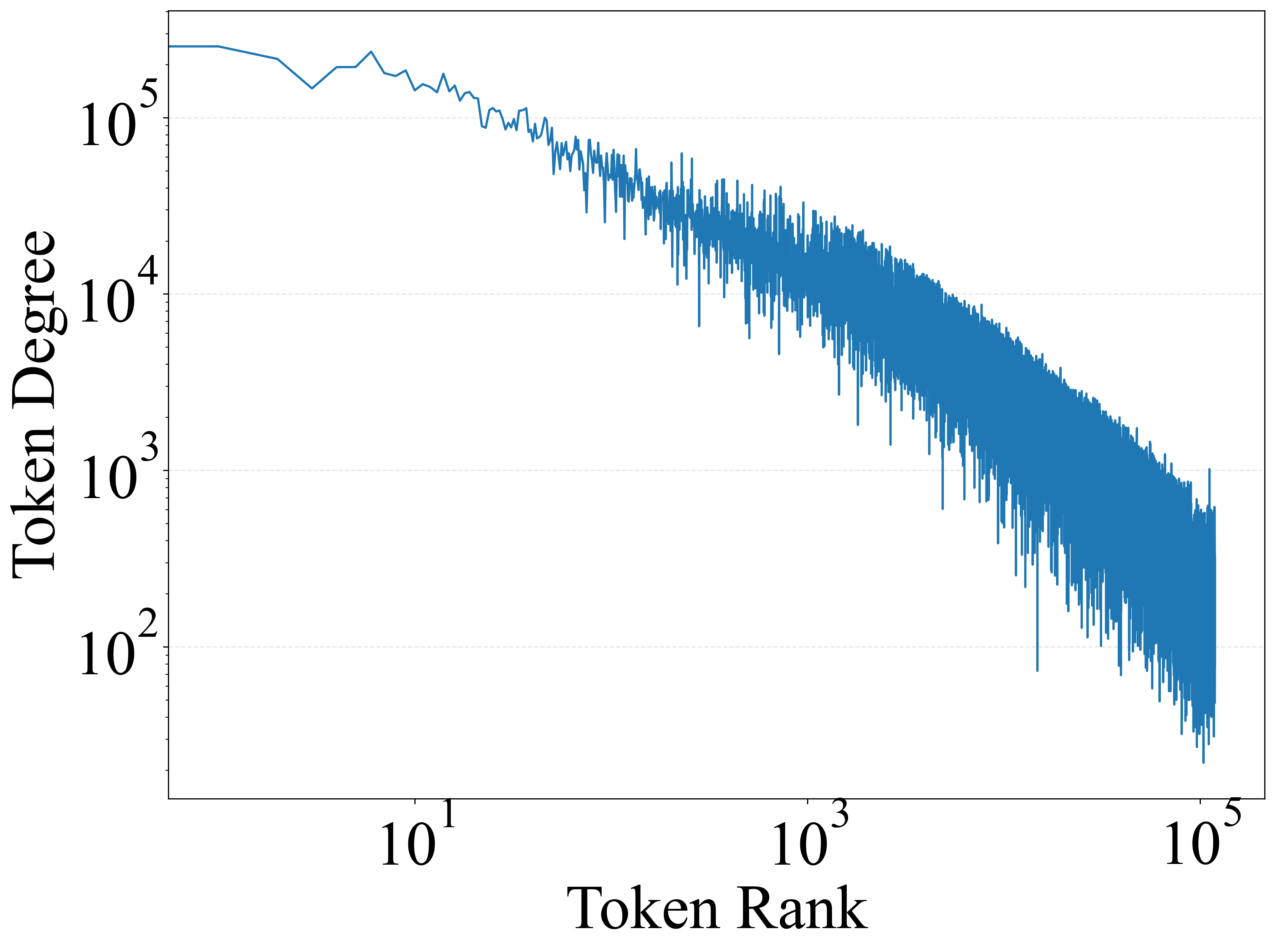}
        \caption{Layer 9}
        \label{fig:sub1}
    \end{subfigure}
    \hfill
    \begin{subfigure}[b]{0.32\textwidth}
        \centering
        \includegraphics[width=\textwidth]{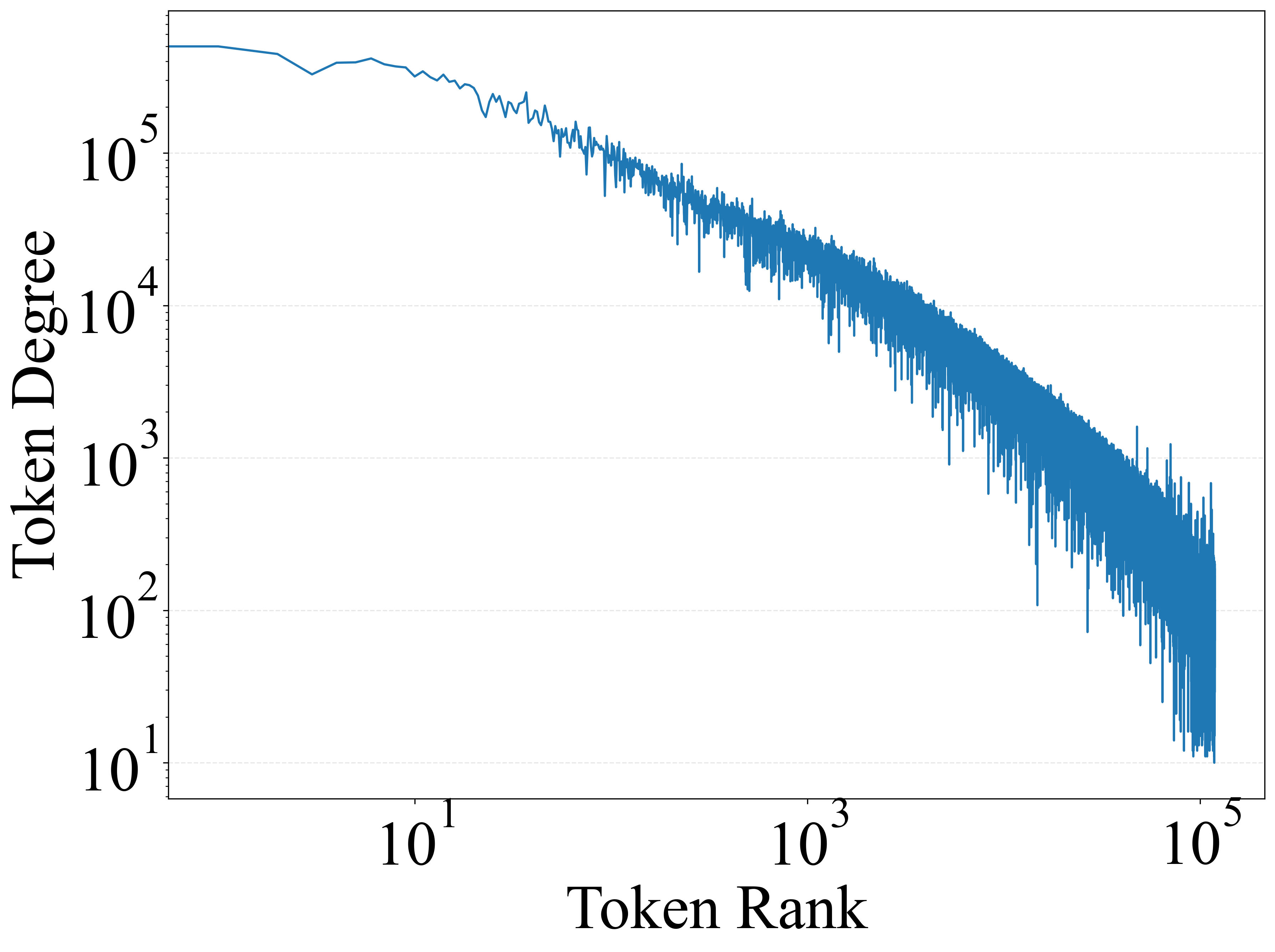}
        \caption{Layer 20}
        \label{fig:sub2}
    \end{subfigure}
    \hfill
    \begin{subfigure}[b]{0.32\textwidth}
        \centering
        \includegraphics[width=\textwidth]{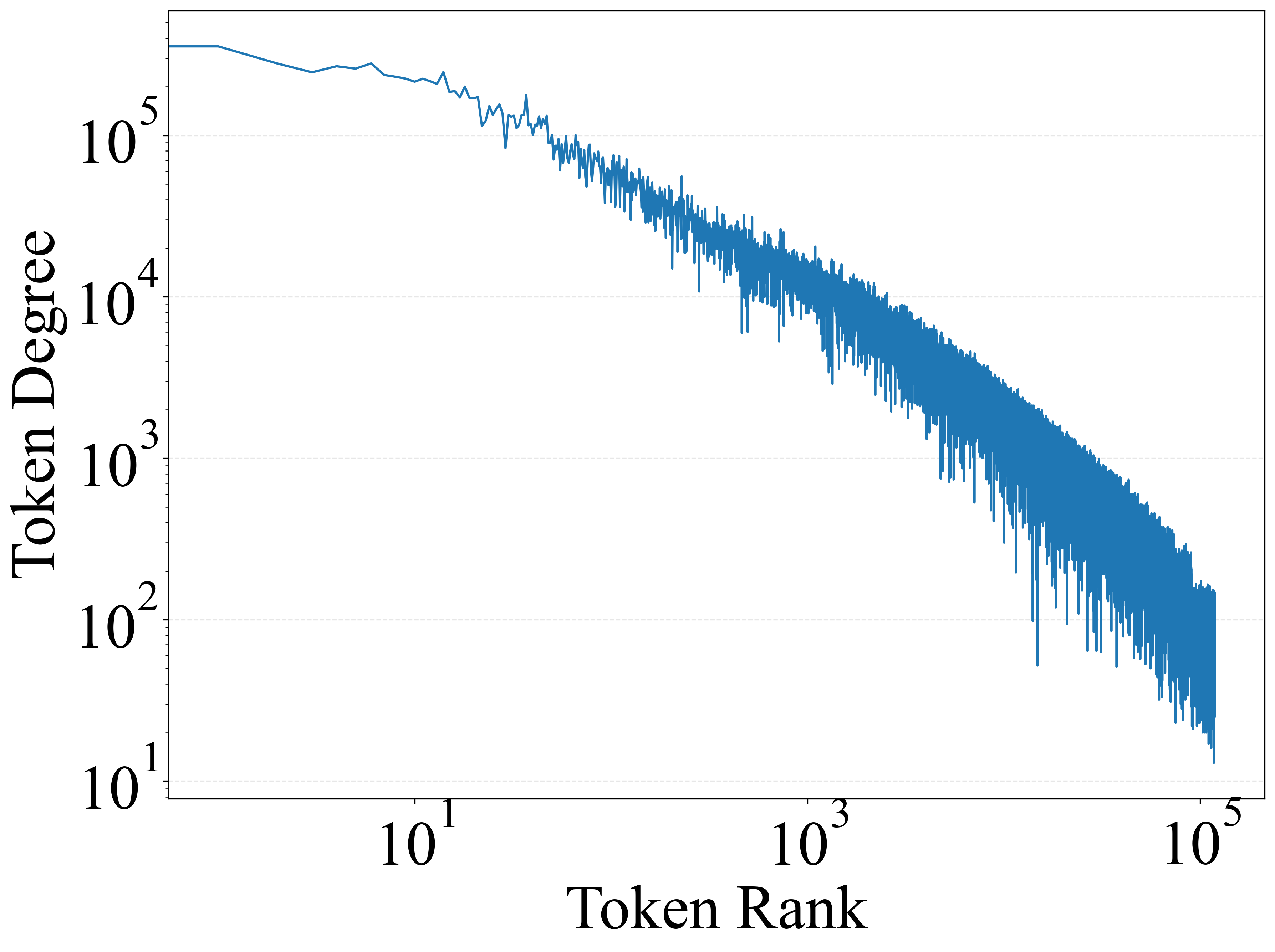}
        \caption{Layer 31}
    \end{subfigure}
    \caption{Token degrees in the token-feature bipartite graph on a log-log scale. Tokens are ranked by frequency from the sampled data.}
    \label{fig:token_degree}
\end{figure}

\begin{table}[h]
\centering
\begin{tabular}{|c|c||c|c|c|}
\hline
\multicolumn{2}{|c||}{Token} & \multicolumn{3}{c|}{Cumulative Feature Coverage} \\
\hline
Rank & Text & Layer 9 & Layer 20 & Layer 31 \\
\hline
1 & . & 23.19\% & 51.32\% & 37.21\% \\
\hline
2 & , & 32.01\% & 62.45\% & 49.78\% \\
\hline
3 & the & 36.88\% & 66.93\% & 55.15\% \\
\hline
4 & \textbackslash n & 39.68\% & 71.30\% & 59.86\% \\
\hline
5 & and & 41.21\% & 71.97 \% & 61.48\% \\
\hline
6 & to & 43.16\% & 73.07 \% & 63.30\% \\
\hline
7 & of & 46.00\% & 74.43 \% & 65.16\% \\
\hline
8 & white space & 47.44\% & 75.70 \% & 67.08\% \\
\hline
9 & a & 47.96\% & 76.12\% & 67.74\% \\
\hline
10 & in & 48.52\% & 76.46\% & 68.27\% \\
\hline
\end{tabular}
\caption{Cumulative feature coverage by top-10 frequent tokens across different layers}
\label{tab:coverage}
\end{table}

Figure~\ref{fig:token_degree} presents the degree of each token in the token-feature bipartite graph. The results reveal that a small set of function tokens can activate most features. Table~\ref{tab:coverage} shows that the top 10 frequent tokens alone account for a substantial proportion of feature activations. In particular, in the middle layer, known to be the most expressive and interpretable~\citep{panickssery2024steeringllama2contrastive,soligo2025convergentlinearrepresentationsemergent,chen2025personavectorsmonitoringcontrolling}, these tokens can activate more than 70\% of the features, demonstrating function tokens's universal access to the feature space. 

\subsection{Feature Reactivation via Function Tokens}\label{sec:feature_compose}

Why can a small number of function tokens activate most features? We hypothesize that function tokens can reactivate the most predictive features, based on preceding contexts.

\begin{figure}[htbp]
    \centering
    \includegraphics[width=1\linewidth]{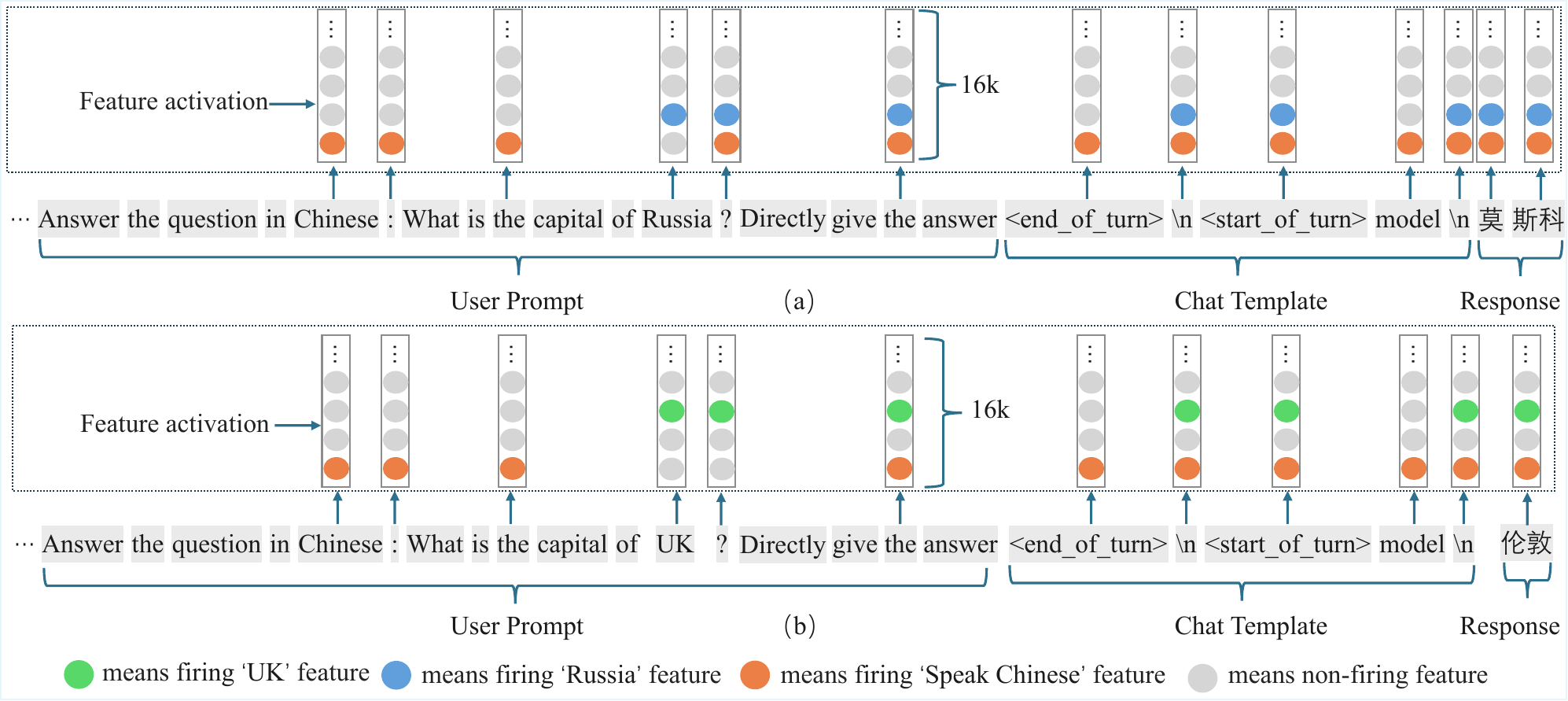}
    \caption{Function tokens can dynamically reactivate predictive features based on different contexts.}
    \label{fig:Steered}
\end{figure}

We design an experiment to examine this hypothesis. First, we identify three interpretable features in Gemma2-9B-it~\citep{gemma_2024}: Feature 15261 corresponds to `Speak Chinese', Feature 9591 corresponds to `Russia', and Feature 13751 corresponds to `UK'. The approach for identifying interpretable features is described in Appendix~\ref{app:model_steer}. We then examine their activations during inference. We employ the following prompt template, wrapping the chat template used in Gemma2-9B-it. 
\tcbset{colback=seedblue!10!white, colframe=seedblue, width=\linewidth, arc=5mm}
\begin{tcolorbox}
\vspace{-5pt}
\subsubsection*{Prompt Template}
\vspace{-5pt}
<bos><start\_of\_turn>user\newline
\{prompt\} Directly answer the question<end\_of\_turn>\newline
<start\_of\_turn>model
\vspace{-5pt}
\end{tcolorbox}
We evaluate the following two prompts and record each token's feature activations, as shown in Figure~\ref{fig:Steered}.
\begin{itemize}
    \item Prompt 1: Answer the question in Chinese: What is the capital of Russia? 
    \item Prompt 2: Answer the question in Chinese: What is the capital of UK? 
\end{itemize}

As shown in Figure~\ref{fig:Steered}, for Prompt 1, the `Speak Chinese' feature is first activated by the token `Chinese', and the `Russia' feature by the token `Russia'. Function tokens such as `:', `the' and `\textbackslash n' serve as conduits for propagating and re-creating these activations. A similar pattern is observed for Prompt 2. Notably, the only difference between the prompts is the replacement of `Russia' with `UK', yet the same function tokens orchestrate different feature combinations, resulting in distinct model outputs.

\begin{figure}[htbp]
    \centering
    \includegraphics[width=1\linewidth, trim=0 150 0 200]{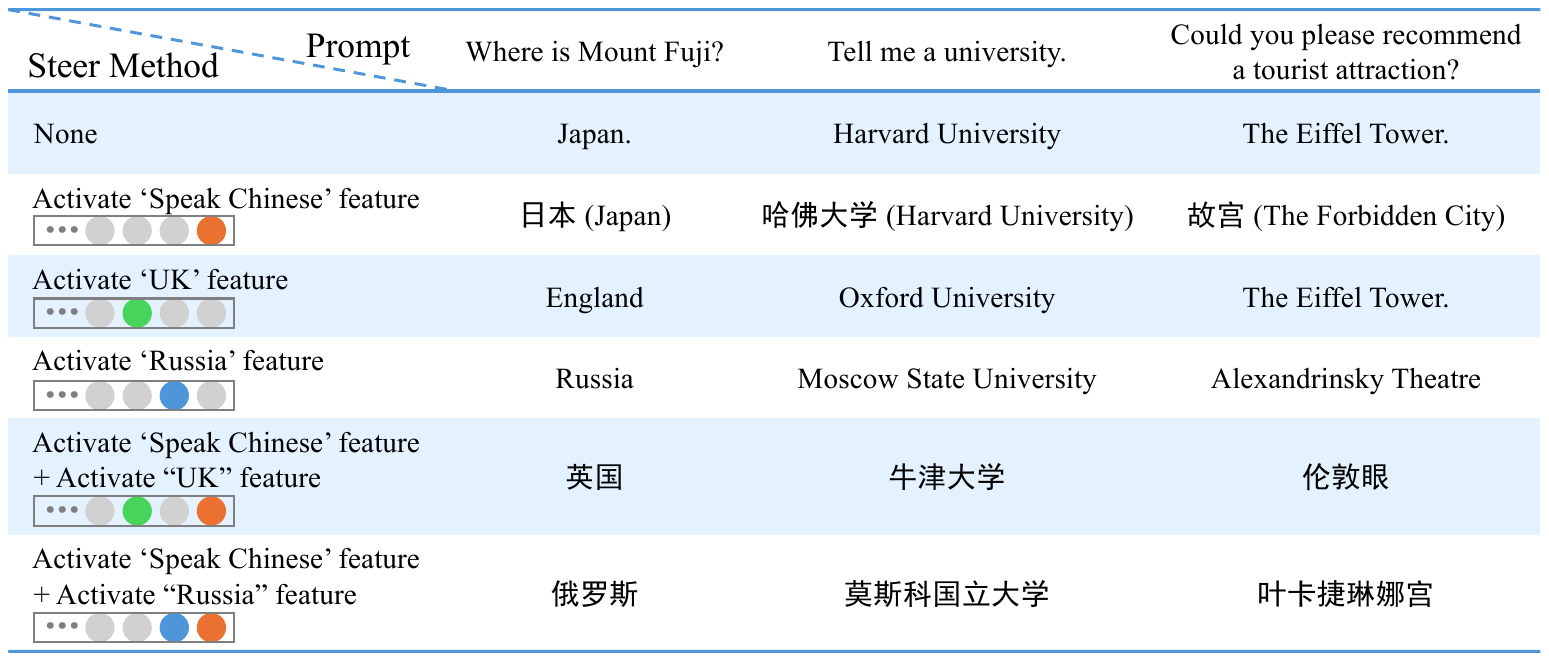}
    \caption{Response of Gemma2-9B-it when editing the activation at the final function token (`\textbackslash n') in the prompt. The Chinese terms shown in the table and their corresponding English translations are: {\begin{CJK*}{UTF8}{gbsn}\raisebox{-0.2ex}{日本}\end{CJK*}} (Japan), {\begin{CJK*}{UTF8}{gbsn}\raisebox{-0.2ex}{哈佛大学}\end{CJK*}} (Harvard University), {\begin{CJK*}{UTF8}{gbsn}\raisebox{-0.2ex}{故宫}\end{CJK*}} (The Forbidden City), {\begin{CJK*}{UTF8}{gbsn}\raisebox{-0.2ex}{英国}\end{CJK*}} (UK), {\begin{CJK*}{UTF8}{gbsn}\raisebox{-0.2ex}{牛津大学}\end{CJK*}} (Oxford University), {\begin{CJK*}{UTF8}{gbsn}\raisebox{-0.2ex}{伦敦眼}\end{CJK*}} (London Eye), {\begin{CJK*}{UTF8}{gbsn}\raisebox{-0.2ex}{俄罗斯}\end{CJK*}} (Russia), {\begin{CJK*}{UTF8}{gbsn}\raisebox{-0.2ex}{莫斯科国立大学}\end{CJK*}} (Moscow State University), and {\begin{CJK*}{UTF8}{gbsn}\raisebox{-0.2ex}{叶卡捷琳娜宫}\end{CJK*}} (Catherine Palace).} 
    \label{fig:casestudy}
\end{figure}

Furthermore, we show that steering activations on function tokens can directly influence model outputs. The steering method is described in Appendix~\S\ref{app:model_steer}. We evaluate this effect using the following prompts:
\begin{itemize}
    \item Prompt 3: Where is Mount Fuji? 
    \item Prompt 4: Tell me a university. 
    \item Prompt 5: Could you recommend a tourist attraction? 
\end{itemize}
As shown in Figure~\ref{fig:casestudy}, features activated by the function token are predictive, driving the subsequent token generation. For Prompt 3, the model normally answers in English (`Japan'). Steering only the activations on the final function token in the prompt (`\textbackslash n') changes the response: activating the `Speak Chinese' feature switches the answer to `{\begin{CJK*}{UTF8}{gbsn}\small\raisebox{-0.2ex}{日本}\end{CJK*}}' (Japan in Chinese), activating the `Russia' feature changes the answer to `Russia', and jointly activating `Speak Chinese' and `UK' features yields `{\begin{CJK*}{UTF8}{gbsn}\small\raisebox{-0.2ex}{英国}\end{CJK*}}' (UK in Chinese). Prompts 4 and 5 exhibit the same behavior, demonstrating that function tokens activate predictive features. For more case studies, see Table~\ref{fig:casestudy_appdix} (\S\ref{app:model_steer}). 

In addition, steering features enable generalized control rather than merely triggering specific word outputs. For example, activating the `Russia' feature can produce contextually appropriate responses, such as `Moscow State University' and `Alexandrinsky Theatre', instead of simply outputting the token `Russia'. This demonstrates that the features encode high-level semantic concepts.




\section{Memory Consolidation through Function Tokens}


We analyze how memory consolidation occurs during pre-training.
We train two models and track their losses on function and content tokens, as well as their feature growth patterns across different training stages. Our key findings are:
\begin{itemize}
    \item As the number of training steps increases, the number of the learned features increases. 
    \item Pre-training initially focuses on learning to predict function tokens.
    \item Subsequently, the optimization process becomes dominated by learning to predict content tokens, especially predicting content tokens that follow function tokens.
\end{itemize}

\subsection{Pre-Training Setup}

We train two models from scratch using the LLaMA-3.1-8B~\citep{grattafiori2024llama3herdmodels} architecture: an 8B model with the originial 32 layers and a 1.5B models with only 2 layers, keeping other components unchanged. We use SlimPajama-627B~\citep{cerebras2023slimpajama} as our pre-training corpus, which is a diverse, high-quality collection of web data that has been carefully deduplicated and filtered. This dataset is well-suited for studying memory consolidation during pre-training. We train for one complete epoch over its 627 billion tokens. We replicate the training hyperparameters of LLaMA-3.1-8B for reproducibility: batch size 1024, max sequence length 4095, AdamW optimizer. The learning rate warm up linearly for 8,000 steps to $8 \times 10^{-5}$, then decays by cosine annealing to $8 \times 10^{-7}$. Training runs on 128 GPUs with 80GB memory each.


\subsection{Memory Consolidation as Feature Expansion}





Due to computational constraints, we perform feature decomposition only on the 1.5B model. To track the number of emergent features during pre-training, we train SAEs on second-layer activations at multiple checkpoints. We use JumpReLU-SAE~\citep{rajamanoharan2024jumpingaheadimprovingreconstruction} with a tanh penalty function~\citep{bloom2024saetrainingcodebase}, wich outperforms alternatives such as TopK-SAE~\citep{gao2025scaling} and Gated-SAE~\citep{rajamanoharan2024improvingdictionarylearninggated}. Training details are in Appendix~\ref{sec:sae_train}.

\begin{figure}[htbp]
\centering
    \begin{subfigure}[b]{0.4\textwidth}
        \centering
        \includegraphics[width=\textwidth]{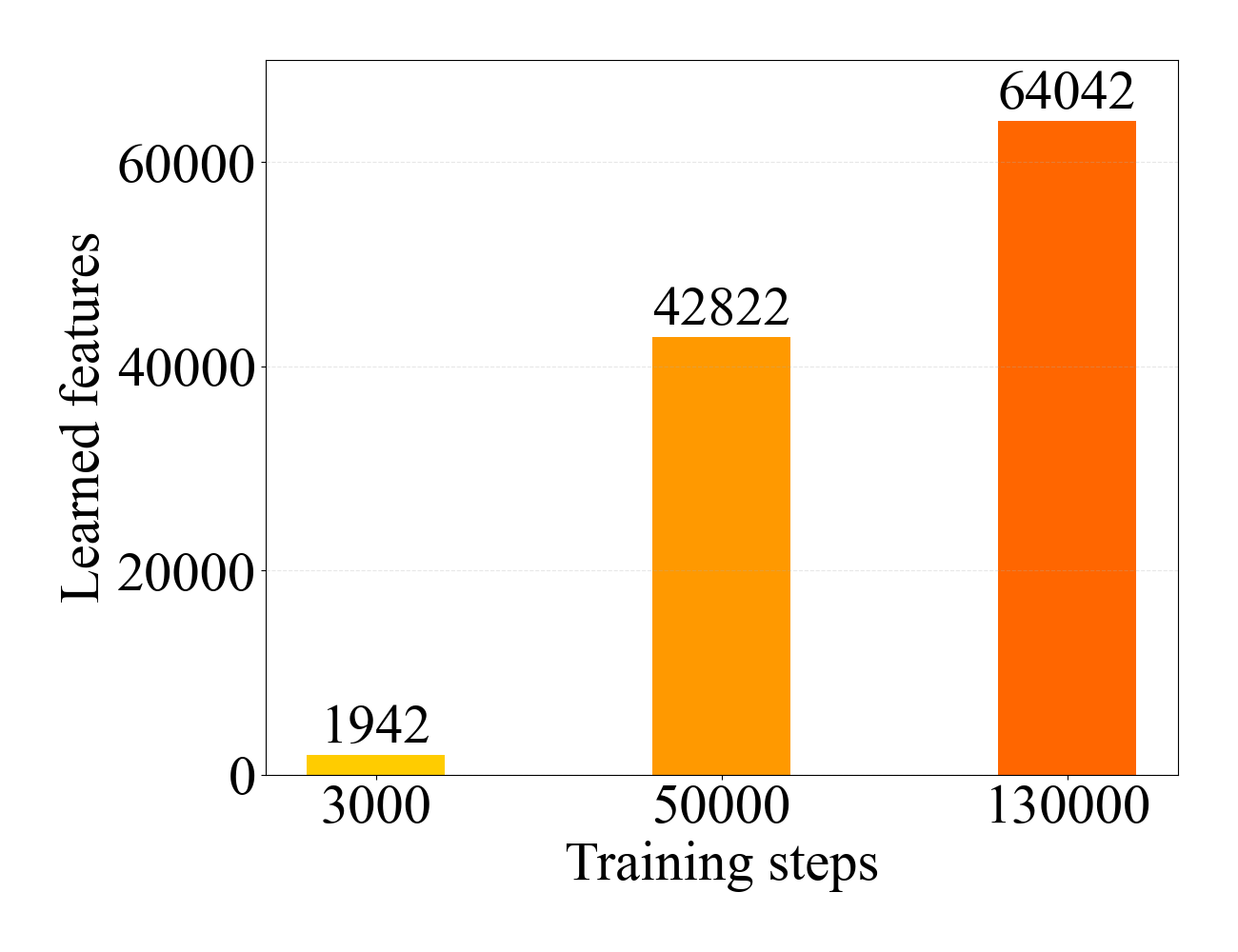}
        \caption{Number of learned features}
        \label{fig:unique_features}
    \end{subfigure}
    \hspace{0.05\textwidth} 
    \begin{subfigure}[b]{0.38\textwidth}
        \centering
        \includegraphics[width=\textwidth]{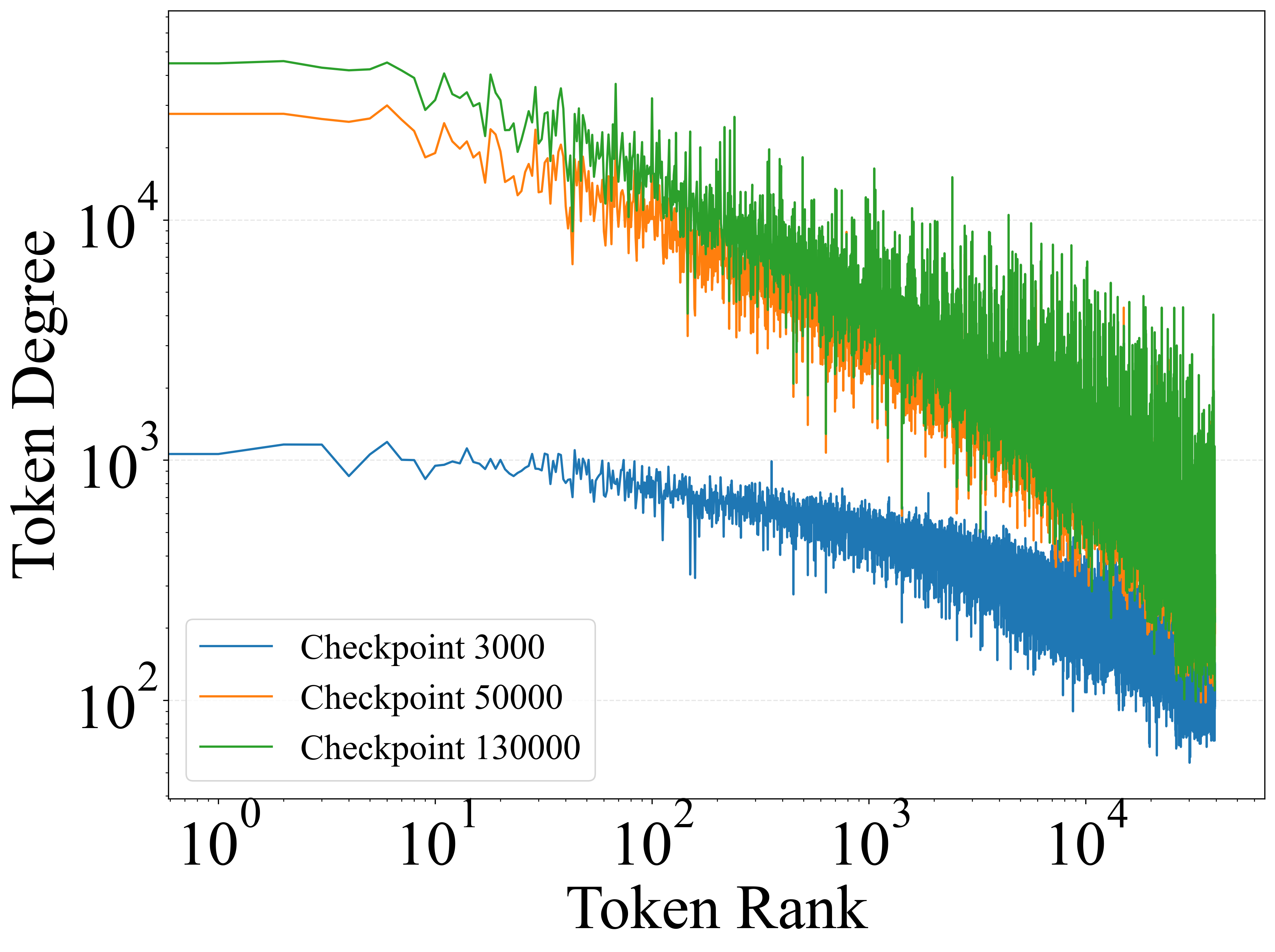}
        \caption{Token degree by checkpoints}
        \label{fig:degree-by-ckpt}
    \end{subfigure}
    \caption{Tracking memory consolidation in relation to feature expansion during pre-training.}
    \label{fig:main}
\end{figure}

We select three representative checkpoints of the pre-training for SAE training: 3000 steps, 50,000 steps and 130,000 steps, corresponding to early, intermediate and late stages of pre-training.  For each checkpoint, we sample text sequences from SlimPajama to obtain 500,000 activations, which are input to the SAE to count the total number of decomposed unique features. As shown in Figure~\ref{fig:unique_features}, the number of features grows substantially over the progress of pre-training, reflecting the model's increasing representational capability and corresponding to memory consolidation.

Using the bipartite graph analysis described in Section~\ref{sec:feature_compose}, we study how token-feature activation evolve. As shown in Figure ~\ref{fig:degree-by-ckpt}, the number of features grows during training, but function tokens consistently activate most features, in contrast with content tokens. This disparity widens over time, as evidenced by the gradually steepening slopes in the graph.

\subsection{Loss on Function and Content Tokens}

\begin{figure}[htbp]
    \centering
    \begin{subfigure}[b]{0.32\textwidth}
        \centering
        \includegraphics[width=\textwidth]{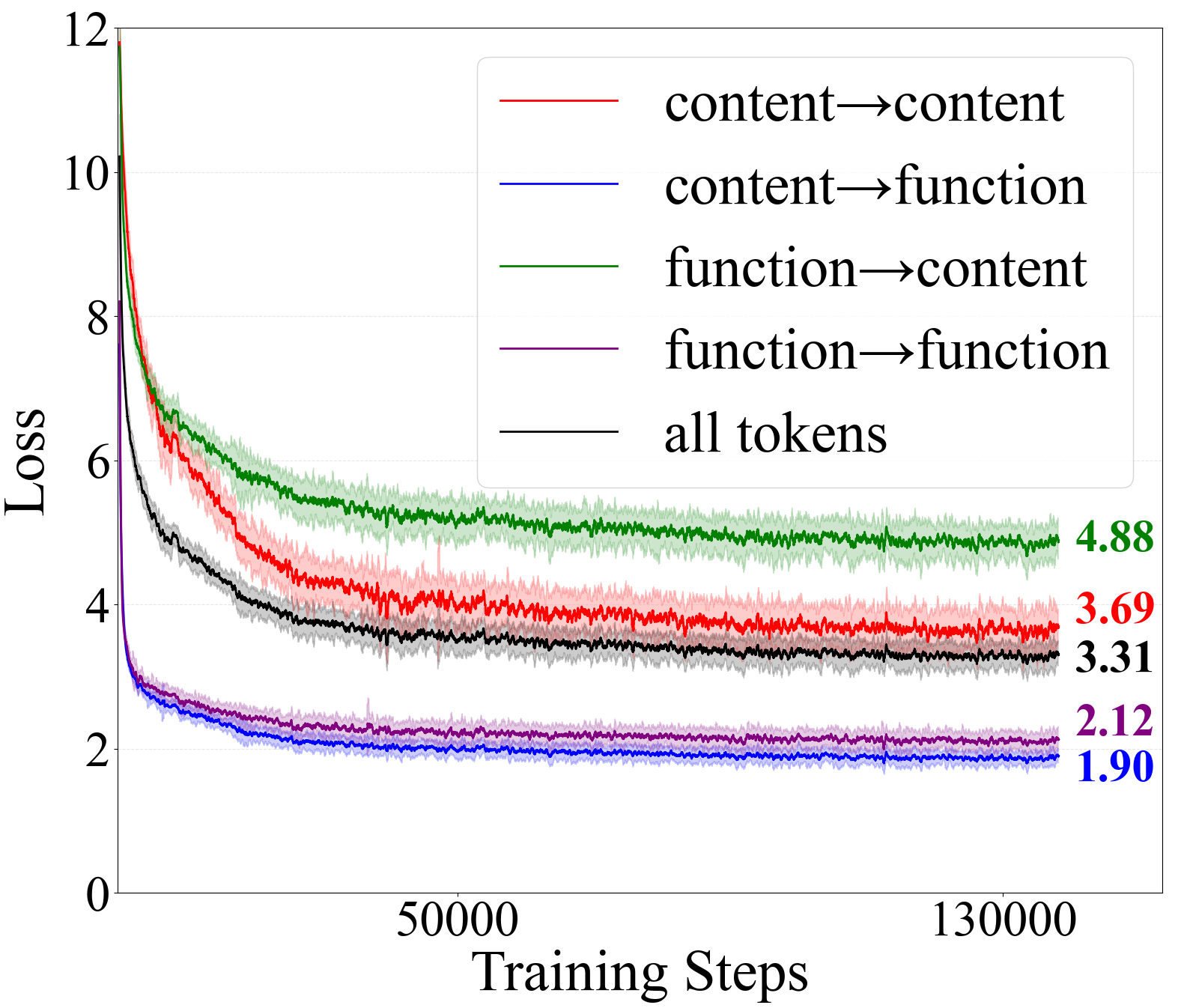}
        \caption{Grouped token loss trajectories during 1.5B model pre-training}
        \label{fig:sub1}
    \end{subfigure}
    \hfill
    \begin{subfigure}[b]{0.32\textwidth}
        \centering
        \includegraphics[width=\textwidth]{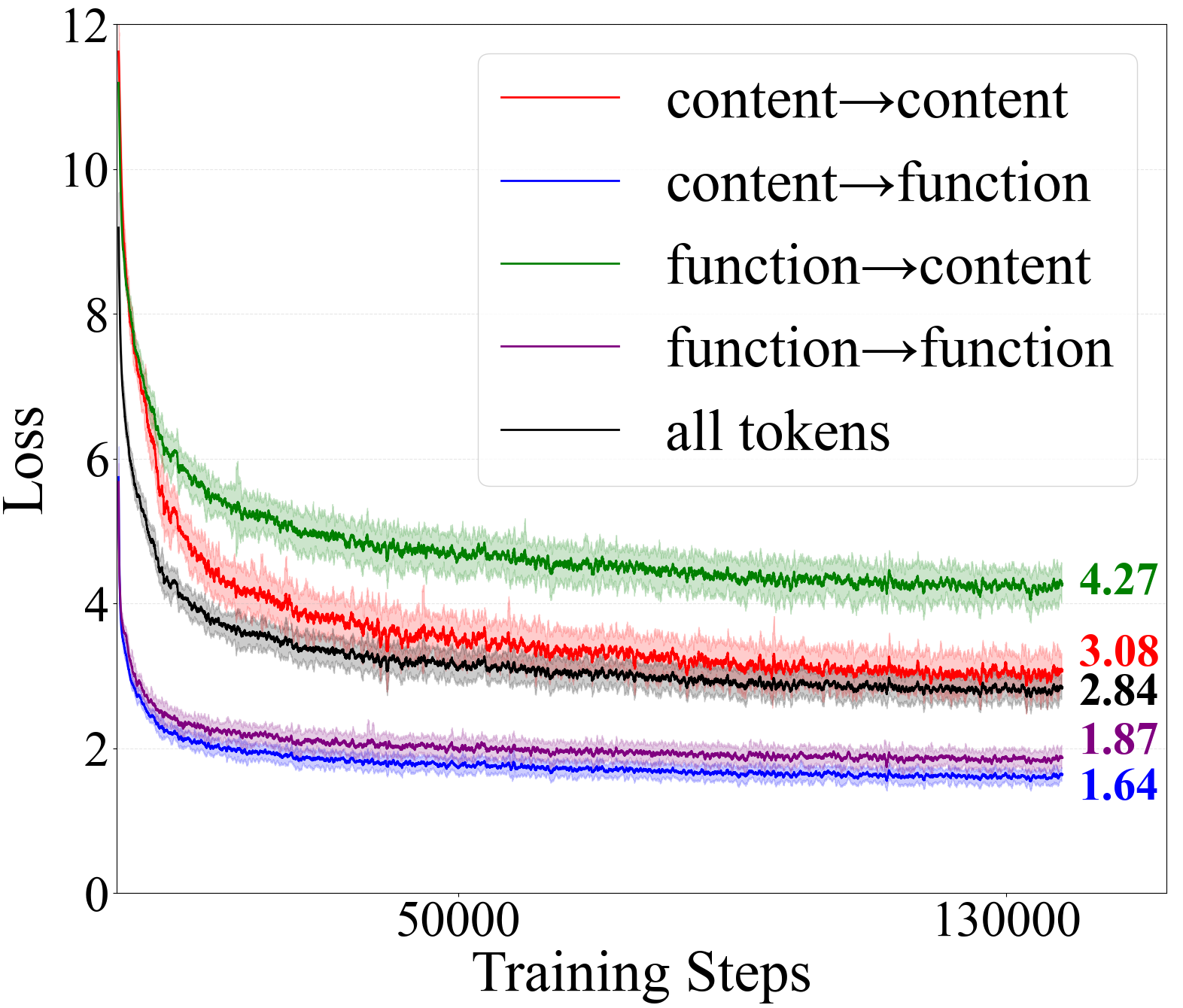}
        \caption{Grouped token loss trajectories during 8B model pre-training}
        \label{fig:sub2}
    \end{subfigure}
    \hfill
    \begin{subfigure}[b]{0.32\textwidth}
        \centering
        \includegraphics[width=\textwidth]{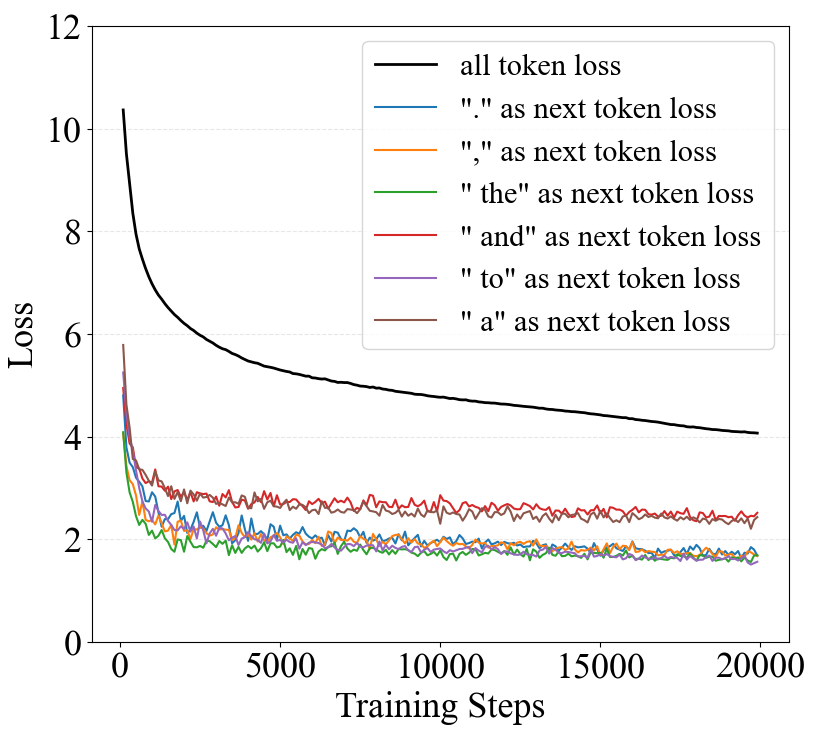}
        \caption{Next-token loss for typical function tokens in 1.5B model pretrain}
        \label{fig:represen}
    \end{subfigure}
    \caption{Pre-training loss curves of different token groups.}
    \label{fig:loss_group}
\end{figure}

To track loss changes for function and content tokens, we categorize next-token prediction of the form $p(\text{next token} | \text{current token}, \text{context})$ into four groups based on whether the current and next tokens are function tokens or content tokens. This yields four distinct categories. For example, $p$(next token = function token $|$ current token = function token, context) is denoted as function→function. The other three categories are defined similarly: function→content, content→function, and content→content. Figure~\ref{fig:loss_group} presents the pre-training loss curves of four groups for both the 1.5B and 8B models, along with the average loss curve across all tokens.
We highlight several key observations.

\noindent{\textbf{Function→Content drives the optimization and memory consolidation.}} Throughout pre-training, the function→content group has the highest loss in both the 1.5B and 8B models, making it the hardest prediction task. As a result, optimization is dominated by this task, which in turn pushes function tokens to develop the capability to reactivate predictive features from context. Furthermore, the feature growth during pre-training likewise primarily driven by function→content prediction.


\noindent{\textbf{Function token prediction is learned faster and more easily.}} For both 1.5B and 8B models, loss decreases more quickly and converge lower when predicting function tokens than content tokens. Function tokens reach very low loss early in training, showing that LLMs first learn to predict function tokens. Figure~\ref{fig:loss_group} plots loss curves of several representative function tokens (`the', `of', and `,') as next tokens to be predicted, alongside the average loss across all tokens, highlighting rapid convergence within the first 3,000 steps. This indicates that the model first learns to generate function tokens before learning to generate more complex token sequences.

\noindent{\textbf{Scaling enhances content token prediction.}} Scaling from 1.5B to 8B parameters yields small loss reductions for content→function group (1.90 to 1.64, $\Delta=0.26$) and function→function group (2.12 to 1.87, $\Delta=0.25$), but much larger loss reductions for function→content group (4.88 to 4.27, $\Delta=0.61$) and content→content group (3.69 to 3.08, $\Delta=0.61$). These results indicate that scaling model size primarily enhances the content token prediction.

\begin{figure}[h]
    \centering
    \includegraphics[width=1\linewidth]{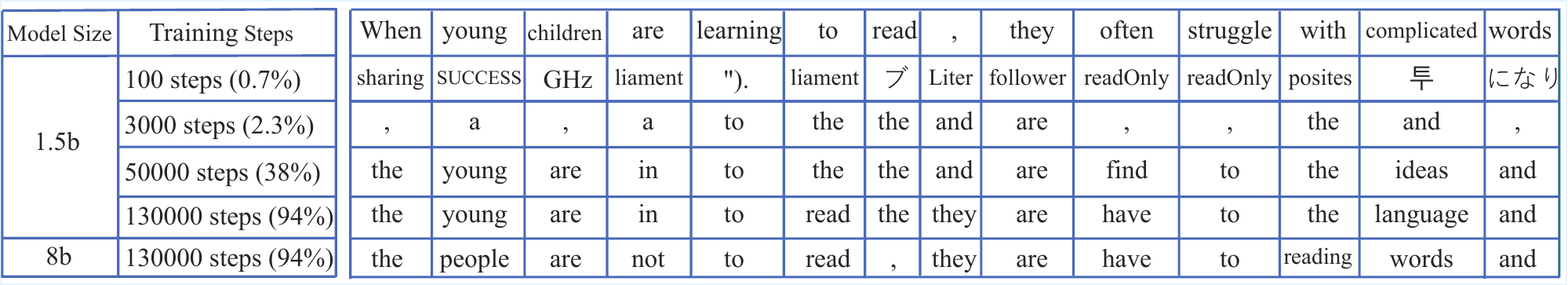}
    \caption{Next token predictions at different training steps. The first row shows the prompt. Each subsequent row shows the next-token predictions conditioned on all preceding tokens. For example, the third column uses `When young' as input, and the fourth uses `When young children' as input.}
    \label{fig:gen_ckpt}
\end{figure}

At last, Figure~\ref{fig:gen_ckpt} provides an example to show how LLM generation evolves during pre-training. At the earliest stage (step 100), generation are random. By step 3,000, the model predicts only function tokens (e.g., `the', `a'). By step 50,000, it generates locally coherent phrases like `learning to' and `to be'. More complex predictions requiring capturing long-range dependencies, emerge at later stages in the 8B model. For instance, correctly predicting the next token after `...struggle with' requires recalling the earlier context 
`learning to read'.

\section{Function Token Hypothesis}

Our experiments suggest that function tokens are crucial for memory consolidation and memory retrieval in LLMs, leading to our Function Token Hypothesis. During inference, function tokens activate the most predictive features from the context to direct the prediction of the next token (memory retrieval). During training, predicting the content token after the function tokens drives parameter updates and feature learning (memory consolidation).

We postulate that the function token hypothesis is the compound result of four factors in LLM training: the training loss (cross entropy loss), learning algorithm (SGD~\citep{ruder2017overviewgradientdescentoptimization} or backpropagation~\citep{rumelhart1986learning}), model architecture (Transformer), and nature of language data.

The training of an LLM is driven by next token prediction.  Maximally reducing the loss for next token prediction means making the prediction as accurate as possible.  (Minimizing the total loss for predicting all next tokens in the training data is equivalent to compressing the training data as compactly as possible.~\citep{delétang2024languagemodelingcompression})  During training, the SGD algorithm always manages to reduce the training loss the most by computing and utilizing the steepest descent.

Each block of the Transformer (decoder-only) consists of a multi-head self-attention layer followed by an FFN layer. Both the self-attention layer and the FFN layer can be viewed as key-value memories, as explained. Their roles, however, are different. The self-attention layer is responsible for producing a new internal vector from all internal vectors in the context (note that compositionality is the key characteristic of language~\citep{chomsky2002syntactic}). The FFN layer is responsible for producing an output vector from the new internal vector. Knowledge is represented as parameters in the FFN layer, and features can be extracted from the output vector.

A natural language text is always segmented by function tokens. From each function token to one of its preceding function tokens, a chunk exists, extending until the beginning of the text. These chunks can represent a phrase, a sentence, or a paragraph, and they are nested. When the LLM’s prediction reaches the token immediately following a function token, this implies the start of predicting the next chunk; the task is far more challenging, as it requires understanding the meaning of the entire context up to that point. This high-challenge prediction compels the LLM to activate the most predictive features in the context during training and reactivate the most predictive features during inference.

Overall, the memory mechanisms of LLMs are extremely complex, due to the complexities of the models and algorithm, as well as the scales of the models and data. Nonetheless, we think that our extensive investigations have convincingly validated the function token hypothesis.

\section{Related Work}
Research on neural memory dates back to the Hopfield network~\citep{hopfield1982neural}, also known as associative memory network, which consolidates memories by adjusting weights between neurons. Hopfield networks have evolved into restricted Boltzmann machines~\citep{fischer2012introduction} and feed-forward networks that utilize key-value memories~\citep{geva2021transformerfeedforwardlayerskeyvalue}. Recent research on superposition~\citep{elhage2022superposition} has shown that it is possible to uncover the features of neural networks such as Transformer, where the number of features is much larger than that of neurons. Through dictionary learning, the superposed activations can be decomposed into monosemantic features. Existing work has demonstrated that such decomposed features can effectively steer model behaviors~\citep{chen2025personavectorsmonitoringcontrolling,panickssery2024steeringllama2contrastive},  by maintaining specific feature activations, controlling access to memories, and directing the model through the generation process.

Existing research on LLMs has identified important patterns involving function tokens. For example, separator tokens produce large activations~\citep{sun2024massiveactivationslargelanguage} and distinct attention weights~\citep{chen2025sepllmacceleratelargelanguage}, enabling efficient KV cache designs retaining only separator caches. The crucial role of ``formatting" in post-training is also widely recognized~\citep{zhou2023lima,ye2025limoreasoning,mamidanna2025onellmssolvemental,li2025llmseasilylearnreason}, a function primarily controlled by tokens such as `\textbackslash n'. Furthermore, recent work on reinforcement learning for reasoning finds that training primarily on high-entropy tokens like  `thus' improves performance~\citep{wang2025beyond}, while Phi-4~\citep{abdin2024phi} identifies `pivot tokens', often following function tokens, as critical for response accuracy.
We argue these are all function tokens, marked by high frequency and diverse contextual usage. This view is supported by previous work demonstrating that the effective learning of function token representations is crucial for overall LLM performance. Building on this, we propose the Function Token Hypothesis and analyze how these tokens drive memory retrieval and consolidation in LLMs.


\section{Conclusion and Open Questions}
In this work, we propose the function token hypothesis: during inference, function tokens activate the most predictive features from context to guide next token prediction. During pre-training, the prediction of content tokens preceding function tokens drives the model to learn and expand its features. Our experiments provide strong evidence for this hypothesis.


In the meantime, our study raises several open questions:
\begin{itemize}
    \item One important question is how function tokens acquire the ability to dynamically activate predictive features, in contrast to content tokens. This capability likely emerges from the interplay of model architecture, data nature, training loss, and learning algorithm during training. Investigating this interaction is essential for a better understanding of the phenomena.

    \item Post-training typically requires only a small number of training steps to achieve substantial improvements in capabilities such as instruction following, chain-of-thought reasoning, and search-agent behavior. Remarkably, training only on function tokens through reinforcement learning can enhance reasoning performance, suggesting that post-training merely activates latent capabilities acquired during pre-training. However, how post-training modifies these activation patterns in function tokens remains an open question.
    \item In our pre-training experiments, we observe that scaling up (more training data, increased computation, and larger model size) reduces loss, accompanied by an increase in the number of learned features. Notably, function tokens consistently activate most features, exhibiting a scale-free property (token-feature degree distribution follows a power law) throughout training. However, the dynamics of feature formation and the underlying reason of this scale-free property remain unclear, and whether these phenomena follow specific principles requires further investigation.
    \item 

    Our case studies confirm existing findings that middle layers offer superior interpretability and steerability. However, the mechanistic explanation for why this steerability is concentrated in middle layers, rather than shallow or deep layers, remains elusive.
    
\end{itemize}
\clearpage

\bibliographystyle{plainnat}
\bibliography{main}

\clearpage

\beginappendix

\section{Steering Method for Large Language Models}\label{app:model_steer}
Given a target trait, our goal is to identify the corresponding feature in Geema2-9B from the SAE decomposition. Specifically, we extract the feature at the last function token in the prompt, which is an newline token. The process involves four steps:

\noindent\textbf{Step 1. Collect contrastive prompts.} We construct two sets of prompts: (i) a single prompt that enforces the target trait, and (ii) 20 prompts that do not. For example, to isolate the `Speak Chinese' feature, we use Prompt 1 (`Answer the question in Chinese: What is the capital of UK?') as the trait-enforcing prompt. This explicitly instructs the model to respond in Chinese. In contrast, Prompt 3 (`Where is Mount Fuji?') is included in the non-trait set, as it contains no language specification and thus defaults to English. The trait-enforcing prompt is used to identify the relevant layer and feature. The non-trait prompts serve as a test set to evaluate steering effectiveness.

\noindent\textbf{Step 2. Identify the most informative layer.} For each layer $l$, we take the activation of the final function token in the trait-enforcing prompt (e.g., Prompt 1 for `Speak Chinese') and denote it as a steer vector~\cite{panickssery2023steering}, $v_l\in\mathbb{R}^d$. We then modify the last function token's activation of each test prompt as $h_l\leftarrow h_l + v_l$, generate responses, and measure the success rate of producing the trait (e.g., `Speak Chinese'). The layer with the highest success rate is chosen as the most informative. For the traits `Speak Chinese', `Russia', and `UK', the most informative layer all correspond to layer 26.

\noindent\textbf{Step 3. Identify the feature.} At the chosen layer, we decompose $v_l$ using SAE. By Equation~\ref{eq:sae_decoder}, $v_l$ can be expressed by
\begin{equation}
    v_l = W_\text{dec}\cdot\textbf{z} + \textbf{b}_\text{dec},
\end{equation}
where $\textbf{z}=(z_1, z_2, \cdots, z_n)^\top$. To locate the trait-specific feature, we rank features by activation strength $z_i$ in descending order. We then apply a binary search to find the smallest $k$ such that activating the top-$k$ features enables the trait, while the top-$(k-1)$ does not. The corresponding steering vector in hidden space is:
\begin{equation}
    v_l^{S_k} = \alpha \cdot W_\text{dec}\sum_{i\in S_k}\textbf{e}_{i}, 
\end{equation}
where $S_k$ is the set of top-$k$ feature IDs, $\textbf{e}_{i}$ is the $i$-th standard basis vector, and $\alpha$ is the steering strength. We apply $h_l\leftarrow h_l + v_l^{S_k}$ on the test set and evaluate the success. For `Speak Chinese', the identified feature is ID 15261 at layer 26; other examples include `Russia' (feature ID 9591, layer 26) and `UK' (feature ID 13751, layer 26).

\noindent\textbf{Step 4. Steering the model.} Once the feature $i$ is identified, the model can be steered with the feature-specific steering vector:
\begin{equation}
    v_l^i = \alpha_i \cdot W_\text{dec}\textbf{e}_i.
\end{equation}
By applying $h_l\leftarrow h_l + v_l^i$ to the last function token of a prompt, we can induce traits such as `Speak Chinese', `Russia' or `UK'.

\section{Additional Case Study}\label{app:case_study}

\begin{figure}[h]
    \centering
    \includegraphics[width=1\linewidth]{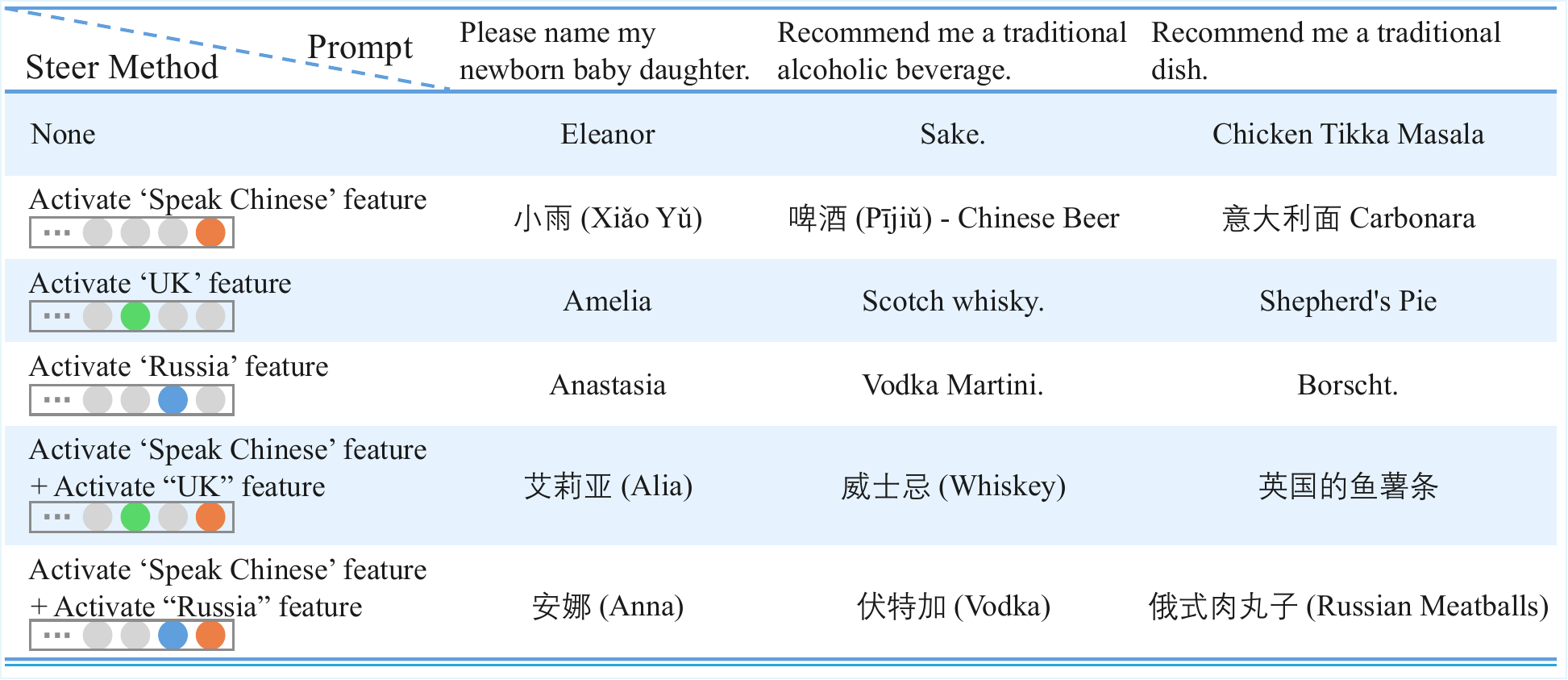}
    \caption{Response of Gemma2-9B-it when editing the activation at the final function token (`\textbackslash n') in the prompt. The Chinese terms shown in the table and their corresponding English translations are: {\begin{CJK*}{UTF8}{gbsn}\raisebox{-0.2ex}{小雨}\end{CJK*}} (Xiaoyu, a common Chinese feminine nickname), {\begin{CJK*}{UTF8}{gbsn}\raisebox{-0.2ex}{啤酒}\end{CJK*}} (beer), {\begin{CJK*}{UTF8}{gbsn}\raisebox{-0.2ex}{意大利面}\end{CJK*}} (Carbonara), {\begin{CJK*}{UTF8}{gbsn}\raisebox{-0.2ex}{艾丽娅}\end{CJK*}} (Alia), {\begin{CJK*}{UTF8}{gbsn}\raisebox{-0.2ex}{威士忌}\end{CJK*}} (Whiskey), {\begin{CJK*}{UTF8}{gbsn}\raisebox{-0.2ex}{英国的鱼薯条}\end{CJK*}} (British fish and chips), {\begin{CJK*}{UTF8}{gbsn}\raisebox{-0.2ex}{安娜}\end{CJK*}} (Anna), {\begin{CJK*}{UTF8}{gbsn}\raisebox{-0.2ex}{伏特加}\end{CJK*}} (Vodka), and {\begin{CJK*}{UTF8}{gbsn}\raisebox{-0.2ex}{俄式肉丸子}\end{CJK*}} (Russian meatballs).}
    \label{fig:casestudy_appdix}
\end{figure}

We present more interesting examples of steering activations on function tokens, as shown in Figure~\ref{fig:casestudy_appdix}.

\section{SAE Training Details}\label{sec:sae_train}


\begin{figure}[h]
    \centering
    \includegraphics[width=0.5\linewidth]{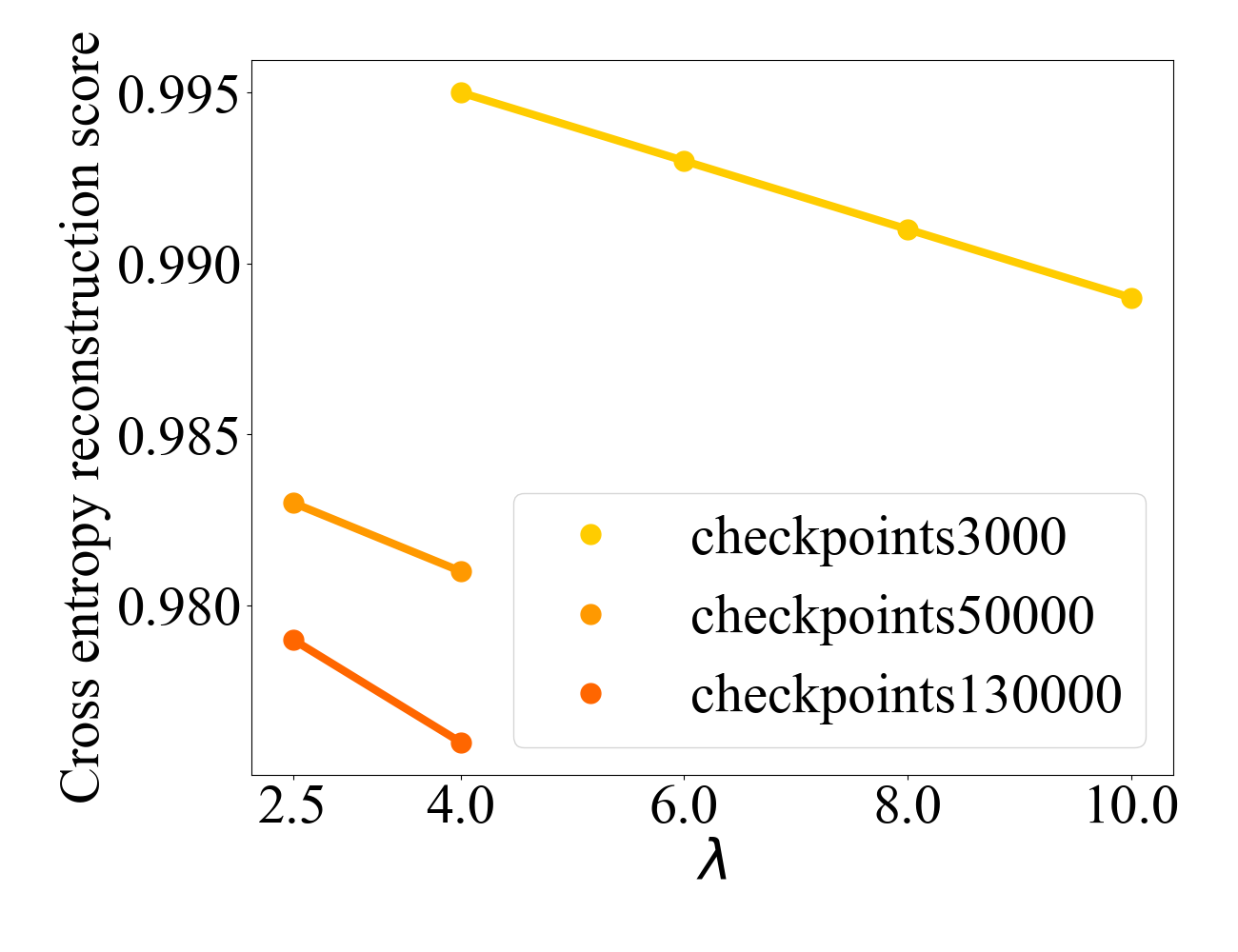}\label{sae_loss}
    \caption{Cross-Entropy reconstruction scores under varying $\lambda$ Values}
    \label{fig:recon}
\end{figure}

Given an activation $\textbf{x}\in\mathbb{R}^d$ from the residual stream with $n$ dimensions, the JumpReLU-SAE comprises an encoder and decoder: \begin{align}
    \mathbf{z} &= \text{JumpReLU}_{\theta} (W_{\text{enc}}\mathbf{x} + \mathbf{b}_{\text{enc}})\\
    \hat{\mathbf{x}} &= W_{\text{dec}} \mathbf{z} + \mathbf{b}_{\text{dec}}\label{eq:sae_decoder}
\end{align}
where $W_\text{enc}\in\mathbb{R}^{n\times d}$, $\textbf{b}_\text{enc}\in\mathbb{R}^n$, $\textbf{b}_\text{dec}\in\mathbb{R}^d$ and $W_\text{dec}\in\mathbb{R}^{d\times n}$. The optimization objective combines reconstruction loss with a $L_0$ sparsity penalty:
\begin{equation}\label{jump_relu_sae_loss}
    \mathcal{L}(\mathbf{x}) = \underbrace{\|\mathbf{x} - \hat{\mathbf{x}}\|_2^2}_{\mathcal{L}_\text{reconstruct}} + \underbrace{\lambda \|\mathbf{z}\|_0}_{\mathcal{L}_\text{sparsity}}
\end{equation}

We train our JumpReLU-SAEs using the open-source library sae\_lens\footnote{https://github.com/jbloomAus/SAELens}~\citep{bloom2024saetrainingcodebase}. Training data consists of one billion activations collected from pre-training dataset~\citep{penedo2024the} with a context size of 1024 tokens. The dictionary width is set to 16 times the activation dimension, resulting in a dictionary size of 65,536. We use a constant learning rate of $1 \times 10^{-5}$.

We adopt default JumpReLU-SAE training setting: batch size 4096 and a dead-feature~\citep{templeton2024scaling} detection window of 1000. For JumpReLU, we set the bandwidth to 0.02 and the initialization threshold to 0.01.

SAE training involves a tradeoff between reconstruction quality and sparsity. To quantify reconstruction quality, we use the cross-entropy reconstruction score~\citep{karvonen2025saebenchcomprehensivebenchmarksparse}, which is defined as $\frac{H_* - H_0}{H_{orig} - H_0}$, where $H_{orig}$ is the cross-entropy loss of the original model for next-token prediction, $H_*$ is the cross-entropy loss after substituting the model activation $x$ with its SAE reconstruction during the forward pass, and $H_0$ is the cross-entropy loss when zero-ablating $x$. The metric ranges from 0 to 1, with higher values indicating more faithful reconstruction.

Using the default $L_0$ penalty coefficient, $\lambda=4$, we find that reconstruction scores varied across early (3,000 steps), intermediate (50,000 steps), and late (130,000 steps) checkpoints, as shown in Figure~\ref{fig:recon}. To make feature counts comparable across these stages, we tuned $\lambda$ to similar reconstruction scores. Specifically, we use $\lambda=10$ for early checkpoint, $\lambda=4$ for the intermediate checkpoint, and $\lambda=2.5$ for the late checkpoint.

From Figure~\ref{fig:recon}, we also observe that, as pre-training progresses, the model's feature representations become increasingly complex and more difficult to decompose.



\section{Function Token List}\label{apppend:func}
Table~\ref{tab:complete_token_stats} presents all function tokens identified in our experiments, ranked by frequency in SlimPajama-627B in descending order. Tokens not appearing in this table are classified as content tokens.
\begin{longtable}{|l|r|r|r|}
\caption{Token statistics with corresponding document coverage, token fractions, and cumulative fractions.} \label{tab:complete_token_stats} \\
\hline
\textbf{Token Text} & \textbf{Document Coverage} & \textbf{Token Fraction} & \textbf{Cumulative Fraction} \\
\hline
\endfirsthead

\multicolumn{4}{c}%
{{\bfseries \tablename\ \thetable{} -- continued from previous page}} \\
\hline
\textbf{Token Text} & \textbf{Document Coverage} & \textbf{Token Fraction} & \textbf{Cumulative Fraction} \\
\hline
\endhead

\hline \multicolumn{4}{|r|}{{Continued on next page}} \\ \hline
\endfoot

\hline
\endlastfoot
, & 95.00\% & 3.60\% & 3.60\% \\
{\_}the & 90.92\% & 3.19\% & 6.79\% \\
. & 95.80\% & 2.31\% & 9.10\% \\
{\_}and & 89.69\% & 1.81\% & 10.91\% \\
{\_}of & 87.59\% & 1.80\% & 12.71\% \\
{\_}to & 88.71\% & 1.68\% & 14.40\% \\
{\_}{\_} & 81.35\% & 1.59\% & 15.99\% \\
{\_}a & 87.62\% & 1.33\% & 17.32\% \\
{\_}in & 86.04\% & 1.16\% & 18.48\% \\
.\textbackslash n & 84.58\% & 0.91\% & 19.39\% \\
{\_}is & 78.90\% & 0.74\% & 20.13\% \\
\textbackslash n & 42.30\% & 0.70\% & 20.84\% \\
{\_}for & 79.82\% & 0.64\% & 21.48\% \\
{\_}that & 67.02\% & 0.62\% & 22.09\% \\
's & 63.02\% & 0.49\% & 22.58\% \\
{\_}on & 72.40\% & 0.47\% & 23.05\% \\
{\_}with & 73.68\% & 0.47\% & 23.52\% \\
{\_}( & 55.05\% & 0.47\% & 23.99\% \\
: & 52.73\% & 0.42\% & 24.41\% \\
{\_}it & 57.50\% & 0.38\% & 24.79\% \\
{\_}I & 37.43\% & 0.38\% & 25.17\% \\
{\_}as & 61.49\% & 0.37\% & 25.54\% \\
{\_}you & 47.06\% & 0.35\% & 25.90\% \\
{\_}be & 60.03\% & 0.33\% & 26.23\% \\
{\_}are & 60.45\% & 0.33\% & 26.56\% \\
{\_}was & 45.51\% & 0.33\% & 26.89\% \\
1 & 40.84\% & 0.30\% & 27.18\% \\
{\_}at & 59.38\% & 0.29\% & 27.48\% \\
{\_}by & 58.44\% & 0.29\% & 27.77\% \\
{\_}`` & 43.01\% & 0.28\% & 28.05\% \\
{\_}The & 55.12\% & 0.28\% & 28.34\% \\
{\_}from & 61.23\% & 0.28\% & 28.62\% \\
) & 44.33\% & 0.28\% & 28.90\% \\
{\_}this & 56.27\% & 0.26\% & 29.16\% \\
{\_}have & 55.12\% & 0.26\% & 29.41\% \\
{\_}or & 50.42\% & 0.25\% & 29.66\% \\
2 & 39.09\% & 0.25\% & 29.91\% \\
- & 38.67\% & 0.24\% & 30.15\% \\
{\_}an & 56.55\% & 0.23\% & 30.38\% \\
0 & 31.70\% & 0.22\% & 30.60\% \\
{\_}not & 46.51\% & 0.21\% & 30.81\% \\
{\_}will & 46.71\% & 0.19\% & 31.00\% \\
{\_}can & 47.99\% & 0.19\% & 31.19\% \\
{\_}has & 49.09\% & 0.19\% & 31.38\% \\
201 & 33.71\% & 0.18\% & 31.56\% \\
{\_}we & 35.13\% & 0.18\% & 31.74\% \\
\textbackslash \textbackslash & 1.30\% & 0.17\% & 31.91\% \\
The & 48.49\% & 0.17\% & 32.08\% \\
{\_}your & 34.99\% & 0.17\% & 32.25\% \\
3 & 35.29\% & 0.17\% & 32.41\% \\
{\_}but & 41.84\% & 0.16\% & 32.57\% \\
{\_}his & 25.09\% & 0.16\% & 32.73\% \\
`` & 34.19\% & 0.16\% & 32.88\% \\
{\_}all & 45.24\% & 0.15\% & 33.04\% \\
{\_}their & 39.27\% & 0.15\% & 33.19\% \\
{\_}he & 23.69\% & 0.15\% & 33.34\% \\
\{ & 1.18\% & 0.15\% & 33.49\% \\
{\_}they & 35.37\% & 0.15\% & 33.64\% \\
't & 33.12\% & 0.15\% & 33.78\% \\
{\_}more & 42.84\% & 0.14\% & 33.93\% \\
{\_}one & 41.94\% & 0.14\% & 34.07\% \\
{\_}which & 40.67\% & 0.14\% & 34.21\% \\
4 & 31.49\% & 0.13\% & 34.34\% \\
5 & 32.71\% & 0.13\% & 34.47\% \\
{\_}\$ & 12.48\% & 0.13\% & 34.61\% \\
{\_}\textbackslash & 0.90\% & 0.13\% & 34.73\% \\
{\_}about & 37.54\% & 0.13\% & 34.86\% \\
{\_}{\_}{\_} & 5.40\% & 0.11\% & 34.97\% \\
; & 21.62\% & 0.11\% & 35.09\% \\
{\_}who & 33.50\% & 0.11\% & 35.20\% \\
{\_}also & 40.22\% & 0.11\% & 35.31\% \\
{\_}our & 30.62\% & 0.11\% & 35.42\% \\
{\_}were & 27.00\% & 0.11\% & 35.53\% \\
{\_}out & 36.49\% & 0.11\% & 35.64\% \\
/ & 20.32\% & 0.11\% & 35.75\% \\
6 & 28.01\% & 0.11\% & 35.86\% \\
{\_}up & 36.43\% & 0.11\% & 35.97\% \\
8 & 28.60\% & 0.11\% & 36.08\% \\
{\_}been & 35.32\% & 0.11\% & 36.18\% \\
{\_}had & 25.51\% & 0.11\% & 36.29\% \\
{\_}if & 30.49\% & 0.10\% & 36.39\% \\
7 & 27.31\% & 0.10\% & 36.50\% \\
{\_}so & 33.25\% & 0.10\% & 36.60\% \\
{\_}my & 20.96\% & 0.10\% & 36.70\% \\
{\_}= & 6.62\% & 0.10\% & 36.80\% \\
{\_}time & 34.79\% & 0.10\% & 36.90\% \\
{\_}her & 15.21\% & 0.10\% & 37.00\% \\
9 & 26.28\% & 0.10\% & 37.10\% \\
{\_}- & 19.91\% & 0.10\% & 37.20\% \\
' & 27.13\% & 0.10\% & 37.30\% \\
s & 28.83\% & 0.09\% & 37.39\% \\
{\_}would & 27.35\% & 0.09\% & 37.49\% \\
{\_}new & 32.43\% & 0.09\% & 37.58\% \\
{\_}when & 32.82\% & 0.09\% & 37.67\% \\
{\_}other & 33.77\% & 0.09\% & 37.76\% \\
{\_}there & 30.15\% & 0.09\% & 37.86\% \\
{\_}A & 28.29\% & 0.09\% & 37.95\% \\
{\_}its & 29.64\% & 0.09\% & 38.04\% \\
{\_}It & 31.56\% & 0.09\% & 38.13\% \\
{\_}like & 30.40\% & 0.09\% & 38.22\% \\
{\_}do & 29.89\% & 0.09\% & 38.31\% \\
{\_}what & 28.23\% & 0.09\% & 38.39\% \\
{\_}{\_}{\_}{\_} & 3.87\% & 0.09\% & 38.48\% \\
{\_}' & 18.94\% & 0.09\% & 38.57\% \\
{\_}into & 31.66\% & 0.09\% & 38.65\% \\
200 & 19.03\% & 0.08\% & 38.74\% \\
\} & 2.01\% & 0.08\% & 38.82\% \\
{\_}than & 30.00\% & 0.08\% & 38.90\% \\
{\_}said & 19.12\% & 0.08\% & 38.98\% \\
{\_}some & 29.97\% & 0.08\% & 39.06\% \\
{\_}them & 27.36\% & 0.08\% & 39.14\% \\
{\_}In & 28.39\% & 0.08\% & 39.22\% \\
{\_}\& & 17.66\% & 0.08\% & 39.30\% \\
{\_}-- & 18.50\% & 0.08\% & 39.38\% \\
{\_}people & 24.05\% & 0.08\% & 39.46\% \\
ing & 29.18\% & 0.08\% & 39.53\% \\
{\_}first & 29.94\% & 0.08\% & 39.61\% \\
)\textbackslash n & 13.24\% & 0.08\% & 39.69\% \\
I & 23.86\% & 0.08\% & 39.76\% \\
? & 24.01\% & 0.08\% & 39.84\% \\
A & 27.74\% & 0.08\% & 39.92\% \\
{\_}just & 27.64\% & 0.07\% & 39.99\% \\
\end{longtable}

\end{document}